\ifdefined\pdfsuppressptexinfo\pdfsuppressptexinfo=-1\fi
\documentclass[10pt]{article}
\usepackage[preprint]{tmlr}
\usepackage{amsmath,amssymb}
\usepackage{booktabs}
\usepackage{longtable}
\usepackage{float}
\usepackage{multirow}
\usepackage{graphicx}
\usepackage{siunitx}
\usepackage{url}
\usepackage{needspace}
\usepackage[hidelinks]{hyperref}
\makeatletter
\newif\ifdeanonymized
\if@accepted\deanonymizedtrue\else\deanonymizedfalse\fi
\makeatother

\newcommand{\R}{\mathbb{R}}
\newcommand{\nGMASE}{\mathrm{nGMASE}}
\newcommand{\nMSIS}{\mathrm{nMSIS}}
\newcommand{\nWQL}{\mathrm{nWQL}}
\newcommand{\nMAD}{\mathrm{nMAD}}

\DeclareMathOperator{\sign}{sign}

\title{TinyCast: Probabilistic Zero-Shot Forecasting \\ with Computed Periodicity}

\author{\name Armin Steinhauser \email armin.steinhauser@raws.at \\
	\addr RAWS Labs}

\begin{document}

\maketitle

	\begin{abstract}
		We introduce TinyCast, an attention-free zero-shot forecaster that emits a predictive
		distribution from \num{146505} parameters, on the premise that at this size the periodic
		structure of a context is worth computing rather than learning.
		A zero-parameter spectral detector supplies the dominant periods, the context is folded on
		their phase, and a dilated convolutional encoder and a block-autoregressive quantile decoder
		model the rest.
		It is smaller than every zero-shot entry on the GIFT-Eval board whose parameter count can be
		established. On probabilistic accuracy it defines the size-accuracy frontier. Among zero-shot
		entries declaring no test-data leakage it is the only one below 1.4\,M parameters that emits a
		predictive distribution, and every entry scoring better carries at least that budget.
		On Chronos-ZS and fev-bench every neural model ahead of it carries at least $28$ times its
		parameters.
		Because the mixing path is convolutions and matrix multiplications only, it exports
		to static INT8 and forecasts end to end on an embedded device without per-signal fitting.
	\end{abstract}

\section{Introduction}

Time series forecasting underlies decision-making in energy, industrial monitoring, logistics and
environmental sensing. For most of its history the field has operated in a one-model-per-dataset
regime: a model is trained on the target series and used to predict its continuation. Time series
foundation models depart from it by pretraining a single model on a large and diverse corpus and
forecasting unseen series zero-shot, without parameter updates \citep{chronos,timesfm,moirai}.
Whether such a model is usable where the data is produced turns on two properties. One is size:
since a single device design is deployed across varying sites, machines and signal types, the
forecaster must fit whatever compute the unit already has. The other is what a forecast carries, since a point
estimate does not tell a consumer how far to trust it, and control loops and alarm thresholds act on
the uncertainty around a forecast \citep{gneiting}. The two have not descended together. Below a
megabyte of parameters, the models that declare no test-data leakage and publish
per-configuration results emit one value per step.

The binding cost of the per-dataset regime is human, and it does not fall as compute becomes
cheaper. A history must be collected and a model fitted, validated and maintained for every signal
at every site, and a fleet of per-signal specialists multiplies acceptance testing and firmware
updates. A per-dataset model can be built with fewer than a thousand parameters \citep{sparsetsf}.
What is missing, however, is generality at that size. The only
alternative that needs no per-signal setup is classical statistics, which fits each series at
inference.

Our question is therefore not how accurate a large forecaster can become, but how small a
probabilistic one can be. We present TinyCast, a time series foundation model built for that budget.
Since at this size any capacity spent rediscovering seasonality is capacity taken from the structure
no fixed computation delivers, TinyCast computes periodicity explicitly, at no parameter cost, and
spends what it learns on the rest.
The model is attention-free: a dilated convolutional encoder, a zero-parameter periodicity detector
computed from a single Fourier transform, and a decoder that emits nine nominal quantile estimates at
any future position, evaluated here to $720$ steps. As every learned operation is a convolution,
a matrix multiplication, a normalization or an elementwise gate, the model stores its weights in
INT8 and needs no operator an embedded runtime does not already ship. It runs in fixed-window
working memory. Trained once, zero-shot, on
GIFT-Eval-Pretrain together with synthetic data, it has \num{146505} parameters.

\needspace{6\baselineskip}
In this work, we make the following contributions:
\begin{itemize}
	\item \textbf{An architecture that computes periodicity rather than learning it.} The context is
	      folded on the phase of the detected periods.
	\item \textbf{Smaller than every zero-shot GIFT-Eval board entry whose parameter count can be
	      established, and, among zero-shot entries declaring no test-data leakage, the only one below
	      1.4\,M parameters that emits a predictive distribution.} Every model in
	      Table~\ref{tab:main} with a better probabilistic score carries at least 1.4\,M
	      parameters. On Chronos-ZS and fev-bench every neural model ahead of it carries at least
	      $28$ times ours. Appendix~\ref{app:census} states which models the comparison admits and
	      why.
	\item \textbf{Controlled experiments that locate the result, and the interventions that failed.}
	      Retraining the shipped budget without the detector costs $0.0071$ in aggregate point
	      accuracy, about eight times the spread across three seeds. Phase binning is the largest
	      single contributor within its family, and the rejected interventions are reported with
	      their measurements (Appendix~\ref{app:negative}).
	\item \textbf{A general zero-shot forecaster running on embedded hardware.} The INT8 core
	      forecasts end to end on the device from one firmware image, with no host, no network and
	      no per-signal fitting. Section~\ref{sec:hardware} gives what the integer path costs.
\end{itemize}

\section{Related work}
\label{sec:related}

TinyCast sits where three lines of work meet: pretrained zero-shot forecasting, attention-free
sequence mixing, and forecasting on constrained hardware. We consider forecasting only, so models
built for classification or representation learning are outside the comparison.

\paragraph{Time series foundation models} Pretrained zero-shot forecasters, from decoder-only and
encoder transformers to state-space encoders and tabular and PFN-based predictors, have closed much of
the gap to per-dataset supervised models without parameter updates
\citep{chronos,chronos2,timesfm,moirai,moirai2,lagllama,flowstate,tabpfnts,tempopfn}. We train on
Chronos's KernelSynth corpus and evaluate against its zero-shot task selection \citep{chronos}.
Comparison rests on benchmarks that publish per-task results for every entrant. GIFT-Eval
\citep{gifteval} spans 97 configurations over seven domains and three forecast terms, and scores
point and probabilistic accuracy separately. Chronos-ZS covers a different 27-dataset selection,
and fev-bench adds covariate-informed and multivariate tasks a univariate model must take one
series at a time. Most
entrants are large and mix sequences by attention or FFT, both hostile to constrained hardware, with
TiRex-2, Chronos-2 and TimesFM-2.5 leading the single-model board entries at the commit we pin
\citep{tirex2,chronos2,timesfm25}. Although linear baselines argue that accuracy need not require
large attention models, DLinear's own GIFT-Eval point accuracy is worse than seasonal naive, at $1.061$, so we take the argument and not the result
\citep{dlinear,gifteval}.

A smaller line of work asks how far the parameter count can fall. TTM \citep{ttm} is a compact mixer
for zero- and few-shot transfer, reserving cross-channel and exogenous modeling for fine-tuning.
Reverso \citep{reverso}, our closest comparator, interleaves long convolutions with DeltaNet
linear-recurrence layers and trains three sizes from 200\,K to 2.6\,M, of which the 550\,K checkpoint
is released. It matches or beats models one to three orders of magnitude larger, and defines the
small end of the frontier we extend. Below 10\,M the field is thin and reaches that scale by making
the learned computation more efficient: dynamic patching in Kairos \citep{kairos}, interleaved block
attention in Xihe-tiny \citep{xihe}, which publishes no per-configuration result and so is not among
our comparators, a cross-variate module in CITRAS-FM \citep{citras}, and FlowState \citep{flowstate}
and Toto-2.0-4m \citep{toto2} in the same band. TinyCast instead takes work out of the learned
computation. Where these models reach periodic structure through attention or patching, we measure
it from the context at no parameter cost. FlowState \citep{flowstate} is the closest precedent,
scaling its state-space encoder by a period it reads from dataset metadata. We detect the period
from the series instead, which is what makes the model usable on a signal it has no metadata for.

\paragraph{Attention-free mixing and explicit periodicity} Our encoder is a dilated causal
convolution stack, the WaveNet primitive \citep{wavenet} in the residual form of the temporal
convolutional network \citep{tcn}, which reaches a long receptive field through local kernels
alone. State-space models \citep{hippo,s4,mamba} and the recurrent xLSTM backbone behind TiRex
\citep{tirex} also stream in constant memory. Only the dilated stack, however, is built entirely
from the kernels an integer runtime ships (Section~\ref{sec:budget}). Measuring
seasonality rather than learning it is equally long established: classical decomposition estimates a
seasonal component directly \citep{stl}, the spectral test we use predates deep forecasting
\citep{fisher1929} and its behavior away from white noise is still being characterized
\citep{quinn2021}, TimesNet folds a series on periods read from its spectrum \citep{timesnet}, and
Autoformer replaces attention with autocorrelation over lag offsets \citep{autoformer}. The closest
work is at our scale: SparseTSF downsamples on a known period below a thousand parameters
\citep{sparsetsf}, CycleNet learns an explicit recurrent cycle and predicts the residual
\citep{cyclenet}, and FITS interpolates in the complex frequency domain at around ten thousand
parameters with edge deployment as its motivation \citep{fits}. LightGTS \citep{lightgts} carries the same lever into a pretrained general forecaster, tokenizing on
an extracted period. All four show that phase-indexed structure is worth its cost at small budgets.
They differ from our setting the same two ways: the periodic hyperparameter, a cycle length for
SparseTSF and CycleNet and a harmonic cutoff for FITS, is supplied per dataset, and the model is
trained on the series it will forecast.

\paragraph{On-device forecasting} TinyML supplies the substrate: efficient kernels
\citep{cmsisnn,tflitemicro}, tiny-model design and search \citep{mcunet,tinyns}, whose objective
presumes a fixed task and so does not transfer when the target signal is unknown at design time, and
on-device time series systems \citep{lighteq}. The forecasters among these are bespoke models trained
for one signal, presupposing that signal known at programming time and a labeled history to train
on. Pretrained models of other modalities have reached this device class, including FEMBA
\citep{femba}, a bidirectional-Mamba electroencephalography model at two-bit weights, and a small
language model generating on an embedded system \citep{deeploy}. We are aware of no report, at the time of
writing, of a general zero-shot probabilistic time series forecaster running on a single-core
Cortex-M-class part with no neural accelerator and no off-chip memory. We place the forecasting core
of such a model on that class of device: one firmware image that forecasts whatever signal the unit
is later pointed at, without per-signal data collection or retraining, and that can serve as an
initialization where a labeled history appears, though we do not test fine-tuning.

\section{Methodology}
\label{sec:method}

\begin{figure}[t]
	\centering
	\includegraphics[width=\linewidth]{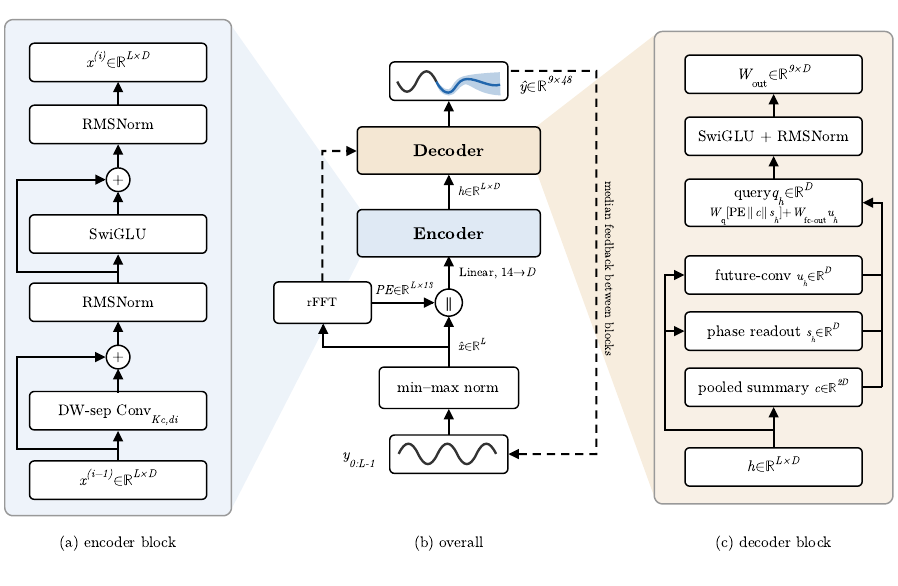}
	\caption{TinyCast architecture. (a)~A single encoder block. (b)~The pipeline for one 48-step
		block: the normalized context and the period-conditioned positional encoding pass through
		the encoder and decoder to nine quantiles at all block positions in parallel; the dashed
		arrow is the median feedback that chains blocks. (c)~The decoder, whose three readouts
		compose the per-horizon query $q_h$.}
	\label{fig:arch}
\end{figure}

A budget of \num{146505} parameters sets the design: whatever the context already exposes to a
fixed computation is computed rather than learned, and the learned parameters are spent on the rest.
Four components follow: a zero-parameter periodicity detector estimates the dominant seasonalities,
a positional encoding turns them into per-position features, a dilated-convolution encoder mixes
the context together with those features, and a decoder queries the encoded context at every future
position (Figure~\ref{fig:arch}). Specific to TinyCast are the
parameterization of the positional encoding by the detected periods, which aligns its phase channels
with the series' own seasonality, and a learned convolutional correction over the horizon axis in the
decoder.

TinyCast is attention-free and forecasts in blocks. Let $y=(y_0,\ldots,y_{L-1})\in\R^L$ denote a
univariate context, $H$ the requested horizon and $p=48$ the fixed block length. One pass maps a
context to nine nominal quantile estimates for one block, at the deciles
$\tau\in\{0.1,0.2,\ldots,0.9\}$, and arbitrary horizons compose this map block-autoregressively:
\begin{equation}
	Q^{(b)} = f_\theta(z^{(b)})\in\R^{9\times p},\qquad
	\hat y=\big[\,Q^{(0)};\dots;Q^{(\lceil H/p\rceil-1)}\,\big]_{:,\,0:H-1},
\end{equation}
where $f_\theta$ is the learned block predictor with parameters $\theta$, $b\geq0$ indexes blocks,
$Q^{(b)}_{0.5,:}$ is the median row, and $z^{(0)}=y$. The next context $z^{(b+1)}$ consists
of the last $L$ values of $[\,z^{(b)};Q^{(b)}_{0.5,:}]$. The median of each completed block becomes
context for the next. The assembled forecast is truncated to the first $H$ steps. The horizon thus
enters only through the number of blocks, and the future offsets $h$ below run over
$\{0,\ldots,p{-}1\}$, position $L{+}h$ being the $(h{+}1)$-th forecast step.

\subsection{Architecture}

The four components appear below in the order a context meets them, after the normalization that
precedes them all.

\paragraph{Normalization} Since series reach the model on arbitrary and unrelated scales, each
context is min-max normalized by its own minimum and maximum. The statistics are detached from the
gradient and their inverse is applied to the output, following Reverso's convention \citep{reverso},
a per-window instance of reversible normalization \citep{revin}. Non-constant
contexts map to $[0,1]$, a bounded range that also suits INT8 quantization
(Appendix~\ref{app:trainsetup}).

\paragraph{Periodicity detection} Because the premise requires the seasonality of an unseen series
to be known before a single parameter is spent on it, the estimate has to come from the context
itself. On the DC-removed context we form the normalized periodogram from one real-valued fast
Fourier transform. We keep the local maxima that pass Fisher's significance test for harmonic analysis
\citep{fisher1929} at $\alpha{=}0.05$, Bonferroni-corrected across bins, and turn the retained bins
into integer periods $p_1,\ldots,p_K$ by rounding. Slots whose peak fails the test are set to zero,
which zeroes their phase channels and collapses their fold to the whole-context average. Four slots ($K{=}4$) let a series hold co-existing cycles, a daily
and a weekly period, for example. Section~\ref{sec:ablations} tests detection against the
metadata-declared period, and Appendix~\ref{app:ablcapacity} tests the cap of four. This is the model's only spectral operation: one $O(L\log L)$ transform per forward pass,
carrying no parameters. It can be re-estimated on every call or cached for slowly varying series
dynamics.

\paragraph{Positional encoding} Because the detected periods are of no use unless the network can
read them at every position, future positions included, they parameterize an encoding evaluated at
any $t\in[0,L{+}H]$ instead of an embedding table. Each period slot contributes a phase pair
$\big(\sin(2\pi t/p_k),\ \cos(2\pi t/p_k)\big)$, zero for a slot that failed the significance test.
Five recency channels carry the signed normalized distance $\Delta(t)$ from the last observed
position: the distance itself, its signed logarithm, which compresses distance symmetrically, and
three exponential decays whose rates form a geometric ladder, the same multi-scale principle as the
encoder's dilation schedule. Like the sinusoidal encoding of the Transformer \citep{transformer} all
channels are deterministic, bounded and free of parameters. Unlike it, the phase frequencies are set
per series by the detected periods, which is what aligns the encoding with the series in front of it.

\paragraph{Encoder} The encoder has to reach across the whole context without a global mixing
operator, for the reason Section~\ref{sec:budget} explains: the operators that mix globally are
the ones the deployment target cannot execute. Dilated causal convolution buys the same reach from
local kernels alone \citep{wavenet}. The value channel, concatenated with the positional encoding
evaluated at the context positions, is projected to width $D{=}64$ and passed through $N{=}10$
blocks. Block $i\in\{1,\dots,N\}$ mixes over time with a depthwise-separable convolution
\citep{mobilenets} of kernel size $K_c{=}3$ and dilation $d_i=2^{\,i-1}$, so the dilations run from
$1$ to $512$ and the receptive field grows to $1+(K_c-1)\sum_{i=1}^{N} d_i = 2047$, one sample short
of the full context, with no downsampling. The context length $L{=}2048$ follows
from what has to be resolvable: a window of that size spans several cycles of even low-frequency
seasonality, the weekly and longer periods in the benchmark's hourly and daily series. Each block
pairs its convolution with a SwiGLU feed-forward \citep{swiglu}, both residual-added and
RMS-normalized \citep{rmsnorm}, and the last encoder block's output is the encoder representation
$h\in\R^{L\times D}$ that the decoder reads. As the padding is causal, every output depends
only on past samples, which is what lets the deployed runtime advance the encoder one position at a
time, and Section~\ref{sec:ablations} finds it free within resolution. Two choices hold the stack
inside the budget. The time mixing is depthwise-separable, and a single SwiGLU is ALBERT-tied
\citep{albert} across all ten blocks, so ten blocks pay for their convolutions and normalizations
but for one feed-forward. Together they remove \num{191680} parameters: the same
architecture without them is at \num{338185} more than twice the deployed size.

\paragraph{Decoder} Since a longer horizon must not cost another encoder pass per step, the decoder
reads one encoding and assembles a query for each future position from three readouts: a pooled
summary, phase binning, and a future-conv correction that convolves across the future positions.

The pooled summary $c=[\bar h\,\|\,h_{L-1}]\in\R^{2D}$, concatenating the mean encoder state and the
last one, is static across the horizon and carries the context-level information. Phase binning is
where the computed periodicity is spent, and the ablations of Section~\ref{sec:ablations} identify it
as the largest single contributor within the architecture family. It restores seasonal structure by
folding the context on its own phase, averaging encoder states that occupy the same position within a
cycle. Every position, future ones
included, receives a phase by the same rule, one of $n_b{=}16$ bins of the period in slot $k$.
Averaging the encoder states that share a bin folds the context into a cycle template of $n_b$
entries per slot. A future position reads each template at its own phase, the $K$ per-slot readouts being mixed by a learned
$W_{\text{phase}}\in\R^{D\times KD}$ into $s_h\in\R^{D}$.
The architecture family's best arm pairs phase binning with a recency gate, a second
readout that bins positions by recency rather than phase and gates its contribution with a learned
sigmoid. The shipped model drops it: removing the gate is part of a reduction that halves the
parameter count without measurable cost (Section~\ref{sec:ablations}).

Summary and phase binning are both horizon-agnostic: each hands every future position the same view
of the context, differing only in which phase that position reads. Neither says anything about how
the series moves near the position being predicted, which is what the third readout adds. It begins
with a draft in value space rather than in encoder space. For each future position the draft is the
average of the past normalized values sharing its phase of the dominant period, a seasonal
continuation that costs no parameters. Each draft is paired with its position's encoding and
projected to width $D$. The $p$ resulting vectors are appended to the last $128$ encoder states, a
six-block causal depthwise-separable network runs over the joined sequence, and its last $p$ outputs
form the correction $u_{0:p-1}\in\R^{p\times D}$. The draft is an input to that network, not a
forecast it emits. Of the three readouts this one costs the most parameters
(Appendix~\ref{app:trainsetup}).

The positional encoding and the three readouts compose the per-horizon query, from which a residual
SwiGLU with RMSNorm and a linear head emit the nine quantiles, de-normalized by the stored context
statistics. Appendix~\ref{app:trainsetup} gives the query and the readouts formally.

\subsection{Training}

Training rolls out four blocks autoregressively under scheduled sampling \citep{scheduledsampling},
so the median feedback the deployed model runs on is seen during training. The base objective is the
nine-quantile pinball loss \citep{koenker1978} plus one term of our own. This gated committing term
of weight $\lambda=0.3$ is active only where repeating the last cycle at the sample's
metadata-derived lag would have beaten the median on the window. It is hinged so that it stops as
soon as the median reaches the copy.
Training is zero-shot with respect to GIFT-Eval: we pretrain on GIFT-Eval-Pretrain
\citep{gifteval}\footnote{\ifdeanonymized
Weights, the code that trains and runs the model, and the evaluation artifacts are at
\url{https://github.com/raws-labs/tinycast} and \url{https://huggingface.co/raws-labs/tinycast}.
\else
Weights, the code that trains and runs the model, and the evaluation artifacts accompany this
submission as anonymized supplementary material, and are released publicly on acceptance.
\fi
The corpus is GIFT-Eval-Pretrain and Chronos KernelSynth from their publishers plus four synthetic
shards the released recipe regenerates; seed and configuration are in the checkpoint.}
after removing every dataset that overlaps the GIFT-Eval and Chronos-ZS test sets and three
gridded-weather corpora. To that we add Chronos
KernelSynth series \citep{chronos} and four synthetic shards, whose contribution
Section~\ref{sec:ablations} measures. Windows are drawn with sampling
balanced across the benchmark's frequency bands under a per-series cap and augmented with temporal
flips, sign flips, downsampling and mixup \citep{mixup}. The sign flip is what the inference-time
symmetrization relies on. The deployed 14-channel encoder receives
one value and 13 positional channels with no mask input. The released weights are the average of the
last eight checkpoints. Appendix~\ref{app:trainsetup} gives the objective, the corpus and the full
configuration.

\subsection{Inference}
\label{sec:budget}

Since inference mirrors the training rollout, the decoder forecasts in blocks of $48$ steps,
producing all positions of a block in parallel, then appends the block's median to the context,
re-encodes, and decodes the next block. One such pass is a core call, the unit of inference work we report
throughout. Blocks are the unit because within a block nothing is fed back, so all $48$ positions are
produced in one pass, while between blocks the context is updated rather than extrapolated.

As the only time mixing inside the encoder is a causal dilated convolution, a forecast runs in fixed-window
$O(L)$ working memory. Only the last encoder block's output is retained for the decoder's pooled summary and phase fold,
so the encoder is evaluated position by position against a bounded per-layer working set
instead of materializing the full-window intermediate activations.
While this set does not grow as the device runs, a forecast still needs to encode its full context
window, a constraint we return to in Section~\ref{sec:limitations}. Every operation on the mixing path is a local convolution or a matrix multiplication, and that locality makes the learned mixing weights
structurally amenable to per-output-channel INT8. The non-affine operations (RMSNorm, the SiLU gate and
input normalization) remain FP32 islands. That locality also separates the dilated stack from the other attention-free
primitives. A softmax attention matrix grows quadratically and softmax is non-affine. An FFT mixer
needs the whole window at once and maps poorly to integer kernels. A state-space scan, though
memory-bounded, needs a scan kernel an integer runtime does not ship, with outlier activation
channels that complicate post-training quantization \citep{mambaptq}. The dilated stack is the only
one of the four that is bounded in memory and built entirely from kernels an integer runtime ships,
and Section~\ref{sec:exp} measures accuracy under static W8A8.

Two inference-time strategies apply off the device. To enforce the sign equivariance the training
augmentation already rewards, the forecast is averaged with the forecast of the sign-negated input,
at the cost of doubling the core calls \citep{timesfm}.
Where the dominant spectral period of a configuration's own test contexts is a clean integer multiple
$k$ of the canonical samples-per-day cycle for its frequency while the canonical cycle itself is
absent from the spectrum, the context is decimated by $k$ before encoding. The coarse forecast is then
interpolated back to the native horizon, in the spirit of FlowState's seasonality-derived temporal
scaling \citep{flowstate}.

\section{Experiments}
\label{sec:exp}

The experiments have to settle four things: whether a forecaster at this parameter budget is
competitive against the released field, which of the components of Section~\ref{sec:method} carry
the result, what that costs under the integer path the deployment target runs, and whether the
integer-friendly mixing path buys execution on constrained hardware.

\paragraph{Setup} We evaluate zero-shot on GIFT-Eval \citep{gifteval}, a benchmark of 97
configurations, each one dataset at one sampling frequency and one of three forecast terms,
spanning seven domains and ten frequencies. Every score is relative to seasonal
naive, the baseline that repeats the value one season earlier. Three metrics measure different
things: the mean absolute scaled error (MASE) for point accuracy, itself the mean absolute error
divided by that of an in-sample one-season-back forecast; the weighted quantile loss (WQL) over the
model's nine quantile levels for probabilistic accuracy; and the mean scaled interval score (MSIS)
for interval quality, which Table~\ref{tab:main} reports and Appendix~\ref{app:extra} analyzes. All three are normalized identically: a configuration's value is divided by
seasonal naive's on that configuration, and the $97$ ratios are combined by geometric mean. We write
these $\nGMASE$, $\nWQL$ and $\nMSIS$, where $1.0$ is parity with seasonal naive
(Appendix~\ref{app:protocol}). GIFT-Eval carries the comparison: its published per-configuration
results let every comparator aggregate be recomputed on identical footing. Chronos-ZS
\citep{chronos} and fev-bench \citep{fevbench} are scored over their own dataset selections, and
neither informed any design decision.

\paragraph{Profiles} Three configurations of the same checkpoint recur below. The host profile is
unquantized and applies both inference-time strategies of Section~\ref{sec:budget}; it is what
Table~\ref{tab:main} reports. The quantized host profile applies the same strategies over static
W8A8. The firmware profile is static W8A8 with neither, the configuration the board of
Section~\ref{sec:hardware} computes. All three are scored on the host, and
Appendix~\ref{app:protocol} gives all eight scored configurations.

\paragraph{Baselines} The census is every zero-shot model up to 10\,M parameters with a public
per-configuration result and no declared test-data leakage (Table~\ref{tab:main}), taken from a
sweep of the full leaderboard. Appendix~\ref{app:census} gives the criteria, the
entries they admitted and excluded, and the source of each parameter count. Four training-free
baselines are scored alongside the census: AutoARIMA, AutoTheta and AutoETS, and FLAIR, which fits a
few dozen coefficients per series at inference. They carry no pretrained parameters, so
Table~\ref{tab:main} places them and the parameter axis of Figure~\ref{fig:frontiers} does not.

\begin{table}[!t]
	\centering\footnotesize
	\caption{Zero-shot GIFT-Eval; lower is better, best per column in \textbf{bold} among the learned
	models. All three metrics are ratios to seasonal naive, which scores $1.000$ on each. Parenthesized values are point errors for models not emitting predictive distributions.}
	\label{tab:main}
	\begin{tabular}{@{}lrrrr@{}}
		\toprule
		Model                   & Params~$\downarrow$ & $\nGMASE\downarrow$ & $\nWQL\downarrow$ & $\nMSIS\downarrow$ \\
		\midrule
		TinyCast (ours)                   & \textbf{146\,K} & 0.774            & 0.545\phantom{)}              & 0.554\phantom{)} \\
		Reverso-Nano \citep{reverso}     & 200\,K  & 0.760               & (0.661)              & (2.035) \\
		Reverso-Small \citep{reverso}    & 550\,K              & 0.726               & (0.626)              & (1.945) \\
		TTM-R3 \citep{ttm}               & 1.4\,M              & 0.724               & 0.520\phantom{)}              & 0.501\phantom{)} \\
		Reverso \citep{reverso}          & 2.6\,M              & \textbf{0.711}      & (0.610)              & (1.905) \\
		Toto-2.0-4m \citep{toto2}        & 4.1\,M              & 0.757               & 0.524\phantom{)}              & \textbf{0.455}\phantom{)} \\
		YingLong-6m \citep{yinglong}     & 7.3\,M              & 0.880               & 0.609\phantom{)}              & 0.534\phantom{)} \\
		FlowState-9.1M \citep{flowstate} & 9.1\,M              & 0.726               & \textbf{0.502}\phantom{)}     & 0.563\phantom{)} \\
		Kairos-10m \citep{kairos}        & 9.9\,M              & 0.753               & 0.554\phantom{)}              & 0.776\phantom{)} \\
		\midrule
		AutoARIMA \citep{autoarima}      & 0\phantom{\,M}               & 1.074               & 0.912\phantom{)}              & 0.948\phantom{)} \\
		AutoTheta \citep{autotheta}      & 0\phantom{\,M}               & 1.090               & 1.244\phantom{)}              & 1.199\phantom{)} \\
		AutoETS \citep{autoets}          & 0\phantom{\,M}               & 1.212               & 7.489\phantom{)}              & 8.635\phantom{)} \\
		FLAIR \citep{flair}              & 0\phantom{\,M}               & 0.838               & 0.587\phantom{)}              & 0.538\phantom{)} \\
		\bottomrule
	\end{tabular}
\end{table}

\subsection{Benchmark results}
\label{sec:results}

\begin{figure}[!tp]
	\centering
	\includegraphics[width=\linewidth]{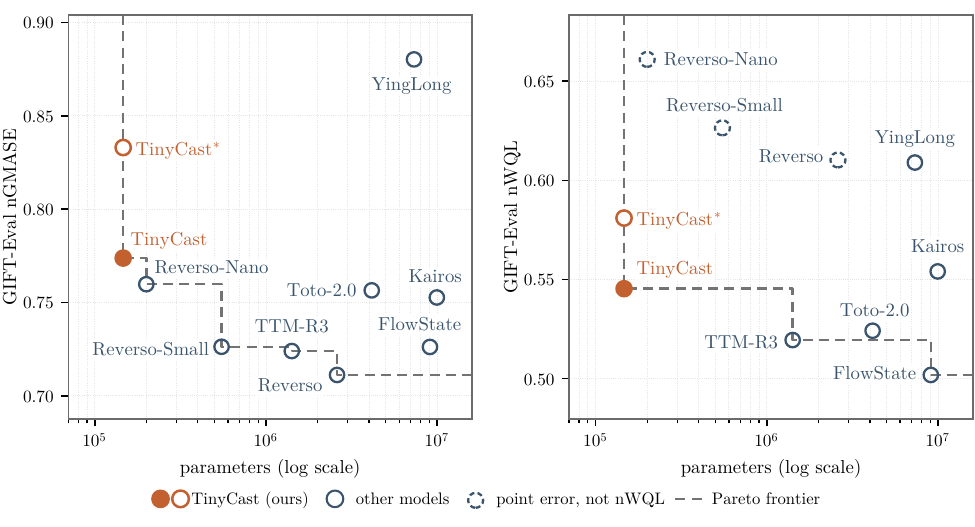}
	\caption{GIFT-Eval point accuracy (left) and $\nWQL$ (right) versus parameter count; lower is
		better. TinyCast$^{*}$ is the same checkpoint at the firmware profile
	(Section~\ref{sec:hardware}).
		Dashed staircase: the Pareto frontier over the census at the host profile, which the
		static-W8A8 point does not enter.}
	\label{fig:frontiers}
\end{figure}

Table~\ref{tab:main} places TinyCast against the census and Figure~\ref{fig:frontiers} plots
$\nGMASE$ and $\nWQL$ against parameter count.

\paragraph{Point accuracy} At \num{146505} parameters TinyCast is the smallest entrant in the census
of Table~\ref{tab:main} and reaches $0.774$ $\nGMASE$, extending the size-accuracy Pareto frontier
(Figure~\ref{fig:frontiers}) to a smaller parameter budget than any model with
published per-configuration results. The next-smallest entrant, Reverso-Nano at 200\,K, scores $0.760$.
Size alone does not order the field: Reverso sets the census best of $0.711$ at 2.6\,M, while
YingLong-6m scores $0.880$ at 7.3\,M. Beyond the census, $33$ entries score lower. Twenty-six of them were verified above the $10$\,M
cut and the board does not size the other seven; the smallest count recorded among them is
TabPFN-TS at $11.1$\,M \citep{tabpfnts}, roughly $75\times$ ours. The statistical baselines sit above parity with seasonal naive, except FLAIR at $0.838$.

\paragraph{Probabilistic accuracy} TinyCast reaches $\nWQL = 0.545$, and every model in
Table~\ref{tab:main} that scores lower carries at least 1.4\,M parameters. TTM-R3 spends 1.4\,M to reach $0.520$ and
FlowState-9.1M spends 9.1\,M to reach $0.502$, roughly $10\times$ and $62\times$ our budget for
$0.026$ and $0.043$ of $\nWQL$.
Beyond the census the floor is the same. Of the $31$ entries that score lower, $25$ were verified
above the cut and six are unsized, and the smallest recorded count is again $11.1$\,M.
Six of the $33$ differences in Table~\ref{tab:main} do not exclude zero once base
datasets are resampled (Appendix~\ref{app:compci}). The interval score is $0.554$
(Table~\ref{tab:main}). Appendix~\ref{app:extra} gives its coverage. FLAIR \citep{flair}, the only training-free baseline that beats seasonal naive, reaches $0.587$ against our $0.545$, and $0.538$ on the interval score against our $0.554$. Appendix~\ref{app:pertask} shows the host-profile forecasts task by task on ten GIFT-Eval tasks.

\paragraph{Chronos-ZS} On the 27-task Chronos zero-shot benchmark, scored against the published
seasonal-naive reference, TinyCast reaches relative MASE $0.880$ and relative WQL $0.722$. The
predictive distribution is again the stronger result and leads every method that needs no training
data. On point accuracy, however, AutoARIMA and AutoTheta are ahead by one to two percent, the only
place across the three benchmarks where a statistical method leads us on point or probabilistic
accuracy. Its aggregates are
not comparable with the GIFT-Eval figures (Appendix~\ref{app:extra}).

\paragraph{fev-bench} Over all 100 tasks \citep{fevbench}, TinyCast reaches relative MASE $0.819$ and
relative WQL $0.658$, ahead of every statistical baseline on both. Every released model that
scores better carries tens of times our parameters, from $28\times$ to $62\times$. On the benchmark's
own primary metric, a scaled quantile loss, the skill score is $0.304$ $[0.247, 0.364]$. Disjointness is not established here: twelve of its tasks name corpus subsets we train on (Appendix~\ref{app:extra}).
TinyCast is univariate and uses neither the covariates that $46$ of the $100$ tasks supply nor the
cross-series structure of the $35$ multivariate ones.

\subsection{Ablations}
\label{sec:ablations}

The design was arrived at through three families of controlled runs, each answering a different
question: which architectural component carries the accuracy, which recipe and inference choices are
worth their cost, and whether the optimization settings sit at their optimum. As each family has its own control and its own training line, deltas are meaningful within a family while absolute
scores are not comparable across families. Every arm is scored on all 97 GIFT-Eval configurations. No family
runs the deployed configuration: the arms carry between roughly 340\,K and 445\,K parameters
against the deployed \num{146505} (Table~\ref{tab:families}). Seven further settings were varied one at a time on an earlier control and are reported in Appendix~\ref{app:ablcapacity}.

\paragraph{What computing the period buys} Since supplying the period costs no parameters,
the design trades structure the model is handed against capacity it would otherwise spend learning
that structure itself. Retrained at
\num{146505} parameters over the full \num{36621} steps with the detector disabled and its readout
inert, the model reaches $0.7814$ $\nGMASE$ and $0.5483$ $\nWQL$ against $0.7743$ and $0.5441$ for
the mean of three independently seeded runs of the shipped recipe. The detector is thus worth $0.0071$
$\nGMASE$ and $0.0042$ $\nWQL$. Every arm here is scored as the eight-checkpoint average the deployed model is reported at. Averaging removes most run-to-run jitter, and the three seeds span $0.0009$ $\nGMASE$,
so on point accuracy the effect is about eight times the
training noise. On $\nWQL$ the seeded spread is $0.0022$, against which the same arm is worth about
twice the noise. Over configurations the contrast does not separate from zero on either metric (Appendix~\ref{app:abldetector}).  Suppressing the detector's output on the trained model costs an order of magnitude more,
which measures how far a model has come to rely on a signal rather than what supplying it is worth.
The retrained arm keeps its phase readout on a degenerate input instead of reallocating the
\num{16448} weights it holds. The contrast therefore measures the detector inside a fixed architecture, and
bounds from above what computing the period buys against a model free to spend those parameters
elsewhere.

What the detector supplies in aggregate is one question; whether its choice of period matters, or
only the shape of its output, is another. To separate the two, three inference-time interventions on the deployed
checkpoint each replace the detected periods and change nothing else: a fixed data-blind set,
the detector's own pooled output applied to the wrong series, and suppression of the output entirely as the matched
control. Against that control neither substitute recovers what suppression costs, and the fixed set
is worse than supplying nothing at all. A well-occupied period the series does not have is an active
harm, where an empty slot set falls back cleanly. What the detector contributes is the correspondence
between a period and the series it was measured from, not the supply of occupied scales.

Within the architecture family phase binning is the largest single contributor, moving $\nGMASE$ by $-0.098$ over
the dilated-convolution base, with its benefit tracking how much periodic structure a configuration
carries. Although it carries $21$\,K parameters more than its control, capacity does not account
for an effect of this size: Appendix~\ref{app:ablcapacity} bounds it two ways. Adding the recency
gate on top gives the family's best result at $-0.113$. However, the gate is cut on the reduction to
the deployed size, in a step that removes three further settings with it and halves the parameter
count at no measurable cost in point or probabilistic accuracy. What the
deployed model takes from this family is phase binning and causal padding. Causal padding, which the streaming mode requires, is free within resolution, with and without phase binning.

\paragraph{Component family} In the component family, whose control is the architecture family's
best arm rebuilt with a nine-quantile head, the
horizon-evolving future-conv correction is the largest single gain at $-0.015$, and combining it with
the synthetic-family blend reaches $-0.026$, the best arm tested. The two are close to additive.
A selection rule with access to metadata does not improve on the raw Fisher pick. Choosing among the
detected periods and the declared canonical and weekly ones by in-context backtest error moves both
aggregates the wrong way, by less than the paired bootstrap can resolve.
Doubling the training rollout from four chunks to eight, the gated committing loss and a MASE-weighted
loss all move the aggregates by less than the paired bootstrap can resolve. The committing loss ships
because it is free and its interval is centered on a gain.

\paragraph{Inference strategies} Since the two inference-time strategies of
Section~\ref{sec:budget} cost compute at every forecast, each has to earn it. Sign symmetrization is
worth $0.0079$ $\nGMASE$ and $0.0061$ $\nWQL$.
Canonical-period alignment is worth $0.0120$ and $0.0112$, but it is narrowly targeted: only two of
the 97 configurations change at all and the remaining $95$ are bit-identical. Both belong to one base
dataset, so resampling base datasets cannot separate the gain from zero
(Appendix~\ref{app:abltables}). The host profile of Table~\ref{tab:main} applies both. The device
runs neither: symmetrization doubles the core calls, and alignment carries a decimation path for a
transform that fires on one base dataset.

\paragraph{Optimization} The learning-rate and budget sweep is flat near its optimum: the chosen
peak beats both neighbors. Raising
the sample budget from 50\,M to the chosen 100\,M buys $0.037$, while the further step to 150\,M, which
the deployed recipe runs, is within resolution of zero. Appendix~\ref{app:abltables} gives every arm, its intervals under both resampling schemes,
and the multiplicity correction.

\subsection{Deployment}
\label{sec:hardware}
Section~\ref{sec:budget} argued deployability from the architecture; the measurements below follow
the integer path from the benchmark to the board.

\paragraph{Quantized accuracy} The first requirement is that the model survive INT8 weights and
activations with frozen scales, checked over the whole benchmark rather than a single input. On all 97
configurations the frozen static-W8A8 path obtains $0.790$ $\nGMASE$ and $0.553$ $\nWQL$ against
$0.774$ and $0.546$ for its matched unquantized reference, degradations of $2.14\%$ and $1.26\%$,
paired per configuration in Appendix~\ref{app:extra}, which also gives the spread.

\paragraph{Firmware configuration} Although the two host-side strategies and
the quantizer are usually reported separately, a device runs all three at once and they
interact. Scored directly over all 97 configurations, with exact static W8A8 and
neither symmetrization nor period alignment, the firmware profile reaches $0.833$ $\nGMASE$ and
$0.581$ $\nWQL$. Composing the three penalties additively predicts $0.810$ and $0.570$. The
shortfall is interaction attributable to no single effect, $0.023$ $[0.015, 0.030]$ on point
accuracy and $0.011$ $[0.005, 0.018]$ on the probabilistic metric under cluster resampling, about
two fifths and a third of the deployment cost.
The interval score degrades about twice as fast as either metric
(Appendix~\ref{app:extra}).  At the firmware
profile the model beats seasonal naive on $72$ of $97$ configurations.

\paragraph{On the board} We deployed the causal TinyCast checkpoint on an STM32H753 (Arm
Cortex-M7) development board as a static-W8A8 graph whose quantization scales are calibrated once
and frozen. Board and host make the same structural decisions. Period detection agrees on every
context, and three independent INT8 backends produce identical outputs on the device
(Appendix~\ref{app:deployment}).

The core call runs from flash within a fixed working set, stable across inputs, so an embedded
scheduler can reserve a slot for it. Nothing else is required, no host, no network and no
per-signal fitting. The alternative a unit would otherwise run is a seasonal-naive ring buffer,
which costs no weights and returns $1.0$ on both metrics with no distribution. For
\SI{138.1}{KiB} of INT8 weights TinyCast returns $0.833$ $\nGMASE$, $0.581$ $\nWQL$ and nine
quantiles. Appendix~\ref{app:deployment} gives the latency and memory record.

		\section{Discussion}
		\label{sec:discussion}

		TinyCast was built on the premise that at a small enough budget, capacity spent
		rediscovering seasonality is capacity unavailable for everything else. The ablations
		support it. Folding the context on the phase of a measured period is the largest single
		contributor within its family, and retraining the shipped budget without the detector is
		worse than keeping it, for parameters the detector never spends. Within that family the
		benefit tracks how much periodic structure a configuration carries, as the premise
		predicts. Between models, however, it does not: against Reverso-Nano, the nearest comparator
		in size, we win a larger share of the configurations the benchmark declares aseasonal than of
		the seasonal ones. The premise is about capacity rather than about seasonality, and a
		returned parameter budget is spent on whatever the series in front of the model contains.

		Two claims follow, and they are not equally strong. Point accuracy at this scale is
		extended by scaling down. Probabilistic accuracy is extended by scaling down and still
		scoring, which is the harder of the two. Since a point estimate does not tell a control
		loop or an alarm threshold how far to trust it, the probabilistic result is also the one
		that decides whether a forecaster this small is usable at all. TinyCast defines that frontier, and every model that scores better on it carries
		parameters by the million.

		The deployment follows from the same budget rather than from a separate engineering
		effort. Since a global mixing operator was out of reach at this parameter count, the
		encoder reaches across its context with dilated convolutions alone, and every learned
		operation is a convolution, a matrix multiplication, a normalization or an elementwise
		gate. That restriction is also what an integer runtime can execute. The architecture the
		parameter budget forced and the architecture that runs on constrained hardware are thus the
		same one, which is why the device needed no redesign, no distillation and no operator an
		embedded runtime does not already ship.

		\paragraph{Limitations}
		\label{sec:limitations}
		A forecast encodes its full context window, so the deployed model repeats that work on
		every call. A per-step streaming variant is possible, using causal normalization and a
		positional encoding invariant to the window advancing, although in a reduced-budget
		ablation it cost accuracy against the windowed model, concentrated in short-context
		configurations. Further limits belong to the method. The model is univariate and reads
		neither covariates nor cross-series structure, so a task that supplies either is forecast
		without it. The model emits no signal when its input leaves the regime its pretraining
		covers, so degradation there is silent. Two properties of the fixed
		computations bound it further: the period is a rounded transform bin, so its resolution
		falls with the ratio of window to period, and the per-window min-max normalization is
		sensitive to a single extreme value.

		Two limits belong to the evidence rather than the model. The ablation
		families run at their own budgets rather than the deployed one, so their deltas bound what
		a component buys at the shipped configuration rather than measuring it. And since GIFT-Eval
		probes informed our architecture, objective and inference choices, Chronos-ZS is the
		untouched test of the process that produced this model.

		\section{Conclusion}

		TinyCast is a probabilistic zero-shot forecaster of \num{146505} parameters. It measures the
		dominant periods of a context spectrally, at no parameter cost, and folds the context on
		their phase, so the learned parameters are left for the structure no fixed computation
		supplies. It is smaller than every zero-shot entry on the GIFT-Eval board whose parameter
		count can be established, and among zero-shot entries declaring no test-data leakage it is the
		only one below 1.4\,M parameters that emits a predictive distribution. What the detector
		contributes is the correspondence between a period and the series it was measured from:
		substituting those periods degrades the model, and supplying well-spread ones degrades it
		further. Because the budget admitted no global mixing operator, every learned operation is a
		convolution, a matrix multiplication, a normalization or an elementwise gate, and that same
		restriction is what lets the model run as a static integer graph on an embedded device.

		Until now, the models at this size that declare no test-data leakage and publish
		per-configuration results have emitted one value per step. A forecaster that also reports how far to trust each step changes what can be
		asked of hardware a deployment already has, since a control loop or an alarm threshold acts
		on an interval rather than on a single number. The mechanism that makes it fit is not
		specific to periodicity. Any structure a fixed computation can supply is capacity returned to
		the learned parameters, and the smaller the budget, the more that return is worth.
		Seasonality is the case where the computation is cheapest and its payoff clearest, and we
		expect the same trade wherever a signal's structure can be measured rather than learned.

\bibliography{references}
\bibliographystyle{tmlr}

\clearpage
\appendix

\section{Architecture, training and inference details}
\label{app:trainsetup}
\suppressfloats[t]

This appendix gives the formal definitions behind Section~\ref{sec:method}. While the body states each
choice and the reason for it, the following section provides the mathematical details.

\subsection{Architecture}

Table~\ref{tab:params} gives the budget component by component. Normalization, the Fisher detector
and the positional encoding carry no parameters at all, so they have no rows.
Roughly $40\%$ of the total sits in the encoder, $30\%$ in the future-conv correction and the
remainder in the decoder readouts.

\begin{figure}[t]
	\centering
	\includegraphics[width=\linewidth]{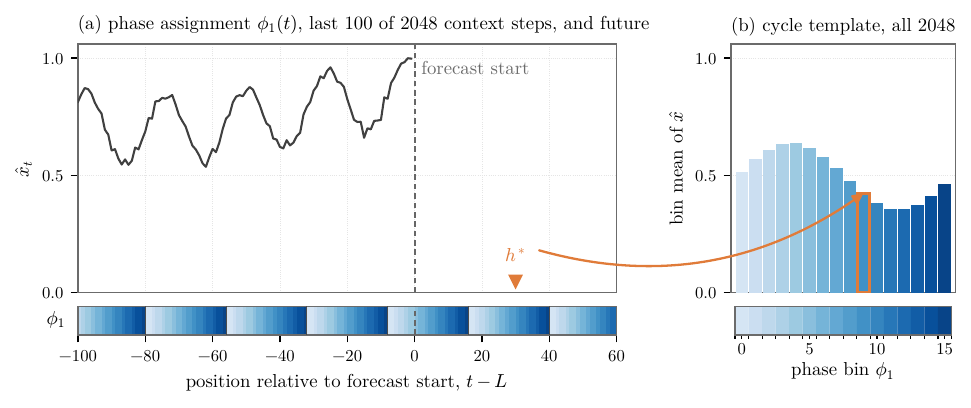}
	\caption{Phase binning of the context with a dominant detected period of $p_1{=}24$. (a)~Context
		values against position relative to the forecast start, $t{-}L$; the stripe below assigns
		every position to one of $n_b{=}16$ phase bins. (b)~Cycle template: the mean of the context values sharing a bin.
		The arrow indicates the phase template selection for the marked future offset $h^{*}$.}
	\label{fig:phasebin}
\end{figure}

\begin{table}[t]
	\centering\footnotesize
	\caption{TinyCast parameter budget: $D{=}64$, $10$ encoder blocks, FFN expansion $1.0$, $K{=}4$
	periods, $n_b{=}16$ phase bins, nine quantile levels.}
	\label{tab:params}
	\begin{tabular}{@{}llr@{}}
		\toprule
		Stage & Module (shape) & Params \\
		\midrule
		input & $\mathrm{Linear}_{\text{in}}$ ($14{\to}64$) & \textbf{960} \\
		\midrule
		encoder & depthwise conv ($10\times(64,1,3){+}\text{bias}$) & \num{2560} \\
		 & pointwise $1{\times}1$ ($10\times(64,64,1){+}\text{bias}$) & \num{41600} \\
		 & shared SwiGLU (up $64{\to}128$, down $64{\to}64$) & \num{12480} \\
		 & RMSNorms ($20\times(64)$) & \num{1280} \\
		 & \textbf{encoder subtotal} & \textbf{57,920} \\
		\midrule
		phase & $W_{\text{phase}}$ ($256{\to}64$) & \textbf{16,448} \\
		query & $W_{\text{q}}$ ($205{\to}64$) & \textbf{13,184} \\
		decoder & SwiGLU $+$ RMSNorm & \textbf{12,544} \\
		\midrule
		future-conv & $W_{\text{fc-in}}$ ($14{\to}64$) & \num{960} \\
		 & depthwise conv ($6\times(64,1,3){+}\text{bias}$) & \num{1536} \\
		 & pointwise $1{\times}1$ ($6\times(64,64,1){+}\text{bias}$) & \num{24960} \\
		 & shared SwiGLU & \num{12480} \\
		 & RMSNorms ($12\times(64)$) & \num{768} \\
		 & $W_{\text{fc-out}}$ ($64{\to}64$, zero-init) & \num{4160} \\
		 & \textbf{future-conv subtotal} & \textbf{44,864} \\
		\midrule
		output & $W_{\text{out}}$ ($64{\to}9$) & \textbf{585} \\
		\midrule
		\textbf{Total} & & \textbf{146,505} \\
		\bottomrule
	\end{tabular}
\end{table}

\paragraph{Normalization} Non-finite inputs are resolved first. In the evaluation path interior
gaps are linearly interpolated, leading and trailing gaps are filled with the nearest observed
value,
any residue is set to zero, and a context shorter than $L$ is left-padded with its first observed
value; the firmware follows the same rule. The 14-channel encoder carries no observed-mask input in
either path. Each context is then min-max
normalized,
\begin{equation}
	\hat x_t=\frac{y_t-y_{\min}}{\max\{\,y_{\max}-y_{\min},\ \varepsilon\,\}},
	\label{eq:norm}
\end{equation}
with $y_{\min}$ and $y_{\max}$ the context minimum and maximum and range clamp
$\varepsilon=10^{-5}$. A constant context maps to zero and remains finite; during training, windows
whose range is at most $10^{-4}$ are excluded from the loss.

\paragraph{Periodicity detector} On the DC-removed context $\tilde y$, zero-padded to a power-of-two
transform length, we form the normalized periodogram via a real-valued fast
Fourier transform (rFFT), threshold it at the large-$N$ Bonferroni approximation to Fisher's level
for harmonic analysis \citep{fisher1929}, and read off integer periods:
\begin{align}
	I[k]     & = \frac{|X[k]|^2}{\sum_{k'\ge 1}|X[k']|^2},\qquad X=\mathrm{rFFT}(\tilde y),
	\label{eq:periodogram}                                                                  \\
	t_\alpha & = \frac{\ln(N_{\text{bins}}/\alpha)}{N_{\text{bins}}},                       \\
	p_k      & = \big\lfloor n_{\text{fft}}/k^{(k)}_{\text{peak}}\big\rceil,\qquad k=1,\dots,K.
\end{align}
The $K{=}4$ largest local maxima of $I$ that exceed $t_\alpha$ with a significance level of
$\alpha{=}0.05$ yield the periods $p_k$, while slots that fail the test are set to zero.
$N_{\text{bins}}$ is the number of positive-frequency bins, $n_{\text{fft}}$ the padded transform
length, $k_{\text{peak}}$ the frequency-bin index of a retained maximum, and $\lfloor\cdot\rceil$
rounds to the nearest integer.

What the test screens out is white noise, not aperiodicity. Against its i.i.d. null it is
calibrated: white-noise surrogates make it declare a period on $4.7\%$ of contexts against a nominal
$5\%$. Autocorrelation alone defeats it. AR(1) surrogates carry no cycle either, yet it declares one
on $98.9\%$ of those with autoregressive coefficient $0.5$ and on essentially all at $0.9$ and
above, so a slot can be filled from a series that has no periodicity.

\paragraph{Positional encoding} The detected periods parameterize an encoding evaluated at any
position $t\in[0,L{+}H]$, with two kinds of channels:
\begin{align}
	\mathrm{PE}(t) & =\big[\,\underbrace{\psi_1(t)\,\|\,\dots\,\|\,\psi_K(t)}_{\text{phase channels}}\,\|\,\underbrace{\rho\big(\Delta(t)\big)}_{\text{recency channels}}\,\big],
	\label{eq:pe}                                                                                                                                                        \\
	\psi_k(t)      & =\begin{cases}
		                  \big(\sin(2\pi t/p_k),\ \cos(2\pi t/p_k)\big) & p_k>0, \\
		                  (0,\,0)                                       & p_k=0,
	                  \end{cases}
	\label{eq:phasepair}                                                                                                                                                 \\
	\rho(\Delta)   & =\big(\,\Delta,\ \sign(\Delta)\log_2\!\big(1+|\Delta|\big),\ e^{-|\Delta|/2},\ e^{-2|\Delta|},\ e^{-8|\Delta|}\,\big),
	\label{eq:recency}
\end{align}
where $\Delta(t)=(t-(L{-}1))/L$ is the signed normalized distance from the last observed position.

\paragraph{Encoder} Each block pairs its dilated separable convolution with a SwiGLU
feed-forward \citep{swiglu}, both residual-added and RMS-normalized \citep{rmsnorm}:
\begin{align}
	x'       & =\mathrm{Conv1d}_{\text{sep},\,d_i}(x^{(i-1)}),                \\
	\tilde x & =\mathrm{RMSNorm}(x^{(i-1)}+x'),                               \\
	x^{(i)}  & =\mathrm{RMSNorm}\big(\tilde x+\mathrm{SwiGLU}(\tilde x)\big),
\end{align}
with causal padding, so length is preserved and each output depends only on past samples;
$x^{(0)}$ is the projected input, and the last encoder block's output is the encoder representation
$h=x^{(N)}\in\R^{L\times D}$ that the decoder reads.

\paragraph{Decoder} Of the three readouts the pooled summary needs no more than its definition.
The other two rest on a phase fold. For period slot $k$,
position $t$ falls into one of $n_b{=}16$ bins,
\begin{equation}
	\phi_k(t) = \Big\lfloor\, n_b\,\frac{t\bmod p_k^+}{p_k^+}\,\Big\rfloor,
	\label{eq:phase}
\end{equation}
with $p_k^+=\max(p_k,1)$ guarding the undetected-period slot. That guard sends every position to bin $0$, so an undetected slot contributes its whole-context average through $W_{\text{phase}}$ rather than nothing. The slot is not masked out. Instead, the network learns what a collapsed slot means from the positional encoding, whose two phase channels for that slot are zero. Because the phase
depends only on the position, future positions receive their bins by the same rule. Averaging the
encoder states that share a bin folds the context into a cycle template,
\begin{equation}
	P^{(k)}(t) = \operatorname*{mean}\big\{\, h_\tau : \tau<L,\ \phi_k(\tau)=\phi_k(t) \,\big\},
	\label{eq:profile}
\end{equation}
so $P^{(k)}\in\R^{n_b\times D}$. A period shorter than $n_b$ occupies only $p_k$ of those rows,
and the rest take the whole-context mean.
Figure~\ref{fig:phasebin} draws the assignment and the template on the fixed test input, a
synthetic window of two sinusoids at periods $24$ and $168$ over a small trend and noise.

The phase readout mixes the per-slot templates,
\begin{equation}
	s_h = W_{\text{phase}}\,\big[\,P^{(1)}(L{+}h)\,\|\,\dots\,\|\,P^{(K)}(L{+}h)\,\big]\in\R^{D},
\end{equation}
and the future-conv draft is the phase average of the normalized values at the dominant slot,
\begin{equation}
	d_h=\operatorname*{mean}\big\{\,\hat x_\tau:\tau<L,\ \phi_1(\tau)=\phi_1(L{+}h)\,\big\}.
\end{equation}

For each future offset $h\in\{0,\ldots,p{-}1\}$ the learned projection $W_{\text{fc-in}}$ maps
$[d_h\,\|\,\mathrm{PE}(L{+}h)]$ to width $D$; these $p$ vectors are appended to the last 128 encoder
states, and the six-block causal depthwise-separable network $\mathcal F$ runs over the joined
sequence,
\begin{equation}
	\tilde u=\mathcal F\!\left(
		[\,h_{L-128:L-1};\,W_{\text{fc-in}}[d_{0:p-1}\,\|\,\mathrm{PE}_{\text{future}}]\,]
	\right),
	\label{eq:futureconv}
\end{equation}
whose last $p$ positions form the correction $u_{0:p-1}\in\R^{p\times D}$. Causal padding prevents
future-label leakage, and the draft $d_h$ is an input to $\mathcal F$ rather than a forecast.

The positional encoding and the three readouts compose the per-horizon query,
\begin{equation}
	q_h = W_{\text{q}}\,[\,\mathrm{PE}(L{+}h)\,\|\,c\,\|\,s_h\,] + W_{\text{fc-out}}\,u_h\in\R^{D},
	\label{eq:query}
\end{equation}
with $W_{\text{q}}$ and $W_{\text{fc-out}}$ learned projections; a residual SwiGLU with RMSNorm and a
linear head then emit the nine quantiles, de-normalized by the stored context statistics.

\paragraph{Remaining specification details} The nine nominal levels are the deciles
$\tau\in\{0.1,0.2,\ldots,0.9\}$ with the median as the fifth. The detector searches only periods in
$[2,\,L/2]$, that is $[2,1024]$ at $L{=}2048$, and only bins that are strict local maxima of $I$
compete, so two adjacent bins can never both be retained. A phase bin that no context position falls into takes the whole-context mean, in the
encoder-state fold and in the value fold of the future-conv draft alike. The future-conv network $\mathcal F$ is
six causal depthwise-separable blocks of kernel $3$ with dilations $1,2,4,8,16,32$, giving it a
receptive field of $127$ over the joined sequence, and one SwiGLU ALBERT-tied across the six, in the
same pattern as the encoder. In the block rollout the value fed into the next context is the raw
$q_{0.5}$ row of the head, taken before any sorting; the nine quantiles are sorted once, on the
assembled forecast, in both the benchmark predictor and the firmware. The normalized head output is
clamped to $[-5,5]$ before de-normalization, matching the clamp the training loss applies, so an
un-penalized overshoot cannot compound through the rollout. Sign symmetrization averages the
forecast with the quantile-reversed forecast of the negated input, since the $\tau$-quantile of
$-y$ is minus the $(1{-}\tau)$-quantile of $y$; on the median the reversal is a no-op. Finally, the decoder is parameterized by future position rather than by a fixed block of output
slots, so the released implementation can also emit an arbitrary horizon in a single pass. Every
score we report comes from the block rollout, and the deployed predictor never takes that
single-pass path.

\subsection{Training}

Training performs a four-block autoregressive rollout. After each
48-step block, its median is fed into the next context with a probability ramped from zero to
$0.5$; otherwise the target block is used. All positions within a block are predicted in parallel.
The future-conv projection $W_{\text{fc-out}}$ is initialized to zero, so the correction starts at
zero and the decoder map reduces to the uncorrected one at initialization.
Table~\ref{tab:training} lists the full training configuration.

\begin{table}[!htb]
	\centering\small
	\caption{Training configuration of the deployed TinyCast.}
	\label{tab:training}
	\begin{tabular}{@{}l p{0.62\linewidth}@{}}
		\toprule
		\multicolumn{2}{@{}l}{\textit{Optimization}}                                                               \\
		Optimizer             & AdamW \citep{adamw}                                                              \\
		Weight decay          & 0.01                                                                             \\
		Gradient clipping     & 1.0                                                                              \\
		Precision             & bf16-mixed                                                                       \\
		Random seed           & 42                                                                               \\
		Total samples         & 150\,M                                                                  \\
		Effective batch size  & 4096                                                                             \\
		Accelerators          & eight RTX 3090, DDP                                                              \\
		Wall clock            & \SI{7.8}{h}, about $62$ accelerator-hours                                        \\
		Final weights         & average of the last eight checkpoints                                            \\
		\midrule
		\multicolumn{2}{@{}l}{\textit{Learning-rate schedule}}                                                     \\
		Shape                 & warmup-stable-decay                                                            \\
		Peak learning rate    & $3\times10^{-3}$                                                                 \\
		Minimum learning rate & $1\times10^{-5}$                                                                 \\
		Warmup fraction       & 5\% (linear to peak)                                                             \\
		Stable fraction       & 60\% (held at peak)                                                              \\
		Decay fraction        & 35\% (to the minimum)                                                            \\
		\midrule
		\multicolumn{2}{@{}l}{\textit{Sequence layout}}                                                            \\
		Context length $L$    & 2048                                                                             \\
		Horizon unit $p$      & 48                                                                               \\
		\midrule
		\multicolumn{2}{@{}l}{\textit{Objective and data}}                                                         \\
		Loss                  & nine-quantile pinball $+$ gated committing term (weight $0.3$)                   \\
		Scheduled sampling    & four blocks; feedback probability ramps to $0.5$                                 \\
		Corpus                & GIFT-Eval-Pretrain $+$ Chronos KernelSynth $+$ four synthetic shards, band-balanced \\
		Augmentations         & temporal flip, sign flip, downsample, mixup; each $p{=}0.5$                      \\
		\bottomrule
	\end{tabular}
\end{table}

\paragraph{Gated committing term} The base objective is the nine-quantile pinball loss. The seasonal
copy $a_h$ repeats the last cycle at the training sample's metadata-derived lag: the rounded
seasonality $\lfloor 24/s\rceil$, clipped to $[2,\,L/2]$, where $s$ is the sample's frequency scale
factor. For the observed target-position set $\mathcal O$, define
\begin{align}
	g &=
	\mathbf 1\!\left\{
		\sum_{h\in\mathcal O}|a_h-y_h|
		<
		\sum_{h\in\mathcal O}|\hat y_{h,0.5}-y_h|
	\right\},\\
	\mathcal L_{\text{commit}}
	&=
	\frac{g}{|\mathcal O|}
	\sum_{h\in\mathcal O}
	\left[
		|\hat y_{h,0.5}-y_h|-|a_h-y_h|
	\right]_+ ,
	\label{eq:loss}
\end{align}
where $[a]_+=\max(a,0)$ and the indicator gate $g$ activates only when the copy beats the median over
the window, so that wherever the median is already the better predictor the term is silent and it
stops as soon as the median reaches the copy. The term is weighted by $\lambda=0.3$ and evaluated
only at the median, while the base loss supervises all nine quantiles.

\paragraph{Corpus} The loader reads six caches: GIFT-Eval-Pretrain, Chronos KernelSynth, and four
synthetic shards of $4096$-step series. GIFT-Eval-Pretrain is the only one of the six caches that carries real series. Before caching it is filtered on two lists, both released with the paper and both matched on the corpus's top-level directory name: a $20$-name benchmark-overlap list, and $66$ gridded-weather names ($33$ CMIP6 slices, $30$ ERA5 years, $3$ WeatherBench splits). Since it is already disjoint from
GIFT-Eval, the first list is there for the secondary benchmark. Of the release's $71$ top-level directories the lists remove two, and the cache keeps $69$. The
run log records $59$ dataset keys across the six caches. Band balancing reweights each real
dataset to the benchmark's own share of configurations per frequency band, $31$ hourly, $30$
sub-hourly, $15$ daily, $8$ weekly, $7$ monthly-and-coarser and $6$ second of the $97$. Synthetic
sources are exempt, and no arm varies the weights. Per epoch a
dataset contributes at most $5{,}000{,}000$ windows and at most $48$ times its series count. Each
window is augmented independently, with probability $0.5$ each, by a temporal flip, a sign flip,
downsampling by an integer stride drawn uniformly from $\{2,3,4\}$, and same-bucket mixup whose
coefficient is drawn from $\mathrm{Beta}(0.2,0.2)$. The scheduled-sampling feedback probability ramps linearly
from $0$ to $0.5$ over the first half of the schedule and holds at $0.5$ thereafter. Checkpoints
are written every $1{,}000$ steps, and the deployed weights are the uniform average of the eight
written at steps $29{,}000$ through $36{,}000$.

\paragraph{Synthetic shards} A note written before the shards were generated, together with the released generator, records what they were built to hold. Three families follow Reverso \citep{reverso}: a Gaussian process over a $38$-kernel bank, trapezoid pulse trains and a trend-seasonal-impulse process, mixed $70/15/15$ at length $4{,}096$, with $62{,}500$ series per shard. On the reduced-budget line that used the same four shards, the synthetic dose was roughly $7\%$ of the mix. The released recipe regenerates
the shards on the pinned stack, so this composition is checkable and the realized dose of the deployed
run is recomputable from the recorded seed and configuration.

\paragraph{Environment} The deployed run trained on eight RTX 3090 GPUs under
distributed data parallelism for \SI{7.8}{h}, about $62$ accelerator-hours, in bf16 mixed precision, with the model compiled by
\texttt{torch.compile} in max-autotune mode.
The versions used were PyTorch $2.10.0$, CUDA
$12.8$ and Python $3.12$, with PyTorch Lightning as the trainer.
Evaluations run through GluonTS $0.15.1$ on Python $3.12$.
The code and weights are released under Apache-2.0, which also covers the two
components they build on: GIFT-Eval's dataset properties and seasonal-naive reference, and the
GluonTS evaluation API. The rollout predictor is adapted from Reverso under MIT, and the
pretraining corpora are used under the terms their publishers set.

\paragraph{Seeds and determinism} Every arm uses seed $42$ except the two repeats of the shipped
recipe at seeds $43$ and $44$, which measure training variance directly. The three runs span
$0.0009$ $\nGMASE$ and $0.0022$ $\nWQL$, which are the spreads we quote. All three are averaged
over the same eight checkpoints, so the figures are comparable; the two repeats alone land $0.0002$
apart on $\nGMASE$, which is one pair rather than a spread. Every other delta in this paper is a single run at seed $42$. The bootstrap
intervals of Appendix~\ref{app:protocol} resample configurations and not runs. Autotuned kernels and the
reduction order of a distributed run leave a rerun free to differ in the last bits. The reported scores are computed under bf16 autocast on an NVIDIA GPU and
depend on that choice at the third decimal: a strict-FP32 CPU rerun reproduces the harness exactly
and still differs from the bf16 record by about $10^{-3}$ on two configurations we checked. The
firmware profile is the exception, its integer arithmetic being exact across three matmul backends and
across boots (Appendix~\ref{app:deployment}).

\subsection{Inference}

Sign symmetrization and canonical-period alignment are both applied at every evaluation of the two
host profiles, and neither is carried by the firmware. Alignment fires only where the dominant
spectral peak of a configuration's own test contexts sits at $k\in[2,16]$ times the canonical
samples-per-day cycle and the canonical cycle is absent from the spectrum. Seven conditions in all
must hold, among them a quorum across the sampled series, a guard that keeps the decimated context
long enough to fill the encoder window and a guard that keeps the horizon long enough to interpolate
back. \texttt{downsample.py} in the released code states all seven with their constants.

\paragraph{Quantile crossing and sorting} The quantile head has no monotonicity constraint, the standard choice for multi-quantile forecasters. Since the population minimizer of each pinball term is the true quantile, the target of training is already non-crossing. Leaving the head unconstrained also keeps the output stage a single INT8-friendly matrix multiplication with an independent median for block feedback. At finite samples, however, crossings still occur, and they are more frequent on the path that ships. On the fixed test input, $21$ of $48$ horizons contain at least one adjacent-quantile inversion in the unquantized path, against $42$ of $48$ under the exact static-W8A8 runtime. Quantization there can move neighboring quantiles onto the same integer level. Sorting is monotone rearrangement, which weakly
improves crossing quantile estimates \citep{chernozhukov2010}, and it is applied in both the
benchmark predictor and the firmware, so the scored object and the emitted object are the same.

\section{Evaluation protocol and comparators}

\subsection{Evaluation protocol}
\label{app:protocol}

Every score comes from the benchmark's own harness, over its own test windows and prediction
lengths, with no added windowing and univariate conversion only where the target is
multivariate. Forecasts are computed under bf16 autocast, and the released code also
carries a strict-FP32 path.

Because the leaderboard's CRPS column reports a weighted quantile loss over each model's finite
quantile grid rather than the continuous ranked probability score, we write $\nWQL$ throughout.

\begin{table}[t]
	\centering\footnotesize
	\caption{The eight scored configurations: one checkpoint, the same 97 GIFT-Eval configurations,
	all scored on the host. Lower is better. ``Exact'' W8A8 calls a host build of the same C
	integer core sources the firmware executes rather than a fake-quantization emulation, and the
	board runs the fidelity chain of Appendix~\ref{app:deployment}, not the benchmark.}
	\label{tab:profiles}
	\begin{tabular}{@{}llccrrr@{}}
		\toprule
		Profile & Arithmetic & Symmetrization & Alignment & $\nGMASE$ & $\nWQL$ & $\nMSIS$ \\
		\midrule
		Host                     & bf16        & yes & yes & 0.7738 & 0.5454 & 0.5541 \\
		No alignment             & bf16        & yes & no  & 0.7858 & 0.5567 & 0.5786 \\
		No symmetrization        & bf16        & no  & yes & 0.7816 & 0.5515 & 0.5638 \\
		Single pass              & bf16        & no  & no  & 0.7935 & 0.5629 & 0.5889 \\
		Quantization reference   & fp32-strict & yes & yes & 0.7736 & 0.5457 & 0.5535 \\
		Single-pass reference    & fp32-strict & no  & no  & 0.7918 & 0.5622 & 0.5865 \\
		Quantized host           & exact W8A8  & yes & yes & 0.7901 & 0.5526 & 0.5632 \\
		Firmware configuration   & exact W8A8  & no  & no  & 0.8328 & 0.5807 & 0.6243 \\
		\bottomrule
	\end{tabular}
\end{table}

Four contrasts in Table~\ref{tab:profiles} carry the costs Section~\ref{sec:exp} reports. Single
pass against host gives the cost of dropping both inference strategies, no symmetrization against
host the cost of dropping symmetrization alone, and no alignment against host the cost of dropping
alignment alone. Quantized host against the quantization reference gives
the quantizer's cost at the strategies the host runs: the reference is that same pass unquantized in
the fp32-strict arithmetic the quantized path uses for its floating-point islands, which is why it
and not the bf16 host row is the denominator. Read against the single-pass reference, which is that
same pairing at the device's own setting, the quantizer costs about twice as much
(Appendix~\ref{app:extra}).

Every comparator number is the model's own published result. For GIFT-Eval we take each
entrant's per-configuration file from the benchmark's public results repository, pinned at commit
\texttt{6fdb10df9c17411f0aef5ff862afbec23627c12f}, and re-aggregate it under the leaderboard's own
rule; for Chronos-ZS and fev-bench we vendor the published per-task files unchanged. Re-aggregating rather than
copying puts every entry on the same seasonal-naive reference over the same $97$ configurations, and
gives the paired intervals of Appendix~\ref{app:compci} per-configuration values to resample. A model that publishes no per-configuration result does not enter the
census, and one that publishes aggregates only, as FLAIR does, is carried at those aggregates. The
fev-bench comparators are each publisher's own run under the harness version current when they
published: $0.8.0$ for FlowState and the statistical baselines, $0.9.0$ for Toto-2.0-4m and $0.6.1$
for CITRAS-FM. The evaluation manifest in the supplementary evidence package records the quantized host
profile, with sign symmetrization and period alignment enabled, and not the firmware profile.

Intervals are percentile bootstraps under one of two schemes: resampling the
$97$ configurations, or clustering, which resamples instead the $28$ base datasets they come from.
Configurations drawn from one dataset are not independent, since the three terms of one
dataset-frequency are the same series at different horizons.

\paragraph{Cost axes} Parameter count sets storage, not arithmetic, and the census separates the two:
the released artifact gives the INT8 weight bytes, whether those bytes alone fit the \SI{2}{MiB}
flash of the device of Section~\ref{sec:hardware}, and how many sequence positions each encoder
evaluates. Four of the nine fit the flash and five do not. On the arithmetic axis the census can only report tokenization. Four of the eight comparators
evaluate their parameters at every one of $2048$ positions as we do, and four are patch-strided,
three of those declaring between $16$ and $128$ tokens per encoder pass. Our core call is $3.61\times10^{8}$ multiply-accumulates at $L{=}2048$, computed from the shapes of Table~\ref{tab:params} and consistent with the $89$ million per second the board sustains over \SI{4.08}{s}. Because two tied SwiGLUs are evaluated at every position, one across the ten encoder blocks and one across the six future-conv blocks, that count is high for our size. It works out to $2465$ multiply-accumulates per parameter per call. For six of the eight comparators the released configuration does not determine a
multiply-accumulate count, so none is estimated.

\subsection{Comparator census}
\label{app:census}

The census fixes the comparison class before any result is read. It admits a model if it
forecasts zero-shot, publishes a per-configuration GIFT-Eval result, declares no test-data
leakage, and holds at most 10\,M parameters, whether or not its checkpoint has been released.
By ``zero-shot'' we mean the model forecasts an evaluation series without fitting to it, which is
the property a device needs. The leaderboard's own \texttt{model\_type} field, however, draws the line differently, reserving \texttt{zero-shot} for models that also do not pretrain on its companion corpus and typing the rest \texttt{pretrained}. Under that field TinyCast and two other census members would be \texttt{pretrained}. No entry of either type whose size can be established falls below our count, the smallest being Reverso-Nano at $200$\,K. Of the rest, $26$ were verified above the $10$\,M cut and $20$ the sweep could not size. The sweep covered every leaderboard entry
at the time of writing; the snapshot is pinned to a single benchmark commit and released with the paper,
and every comparator aggregate in Table~\ref{tab:main} is recomputed from it against the same
seasonal-naive reference, so no number is copied from a publishing paper. AutoARIMA, AutoTheta
and AutoETS enter through each benchmark's own published run of StatsForecast
\citep{statsforecast}.

\paragraph{Excluded, with reasons} The 10\,M cut removes the models the census is not drawn
against, TempoPFN among them: at 38\,M it declares zero-shot and no leakage and re-aggregates on
the snapshot to $0.7875$, $0.5327$ and $0.4910$, behind us on point accuracy and ahead on both
probabilistic metrics. Four sub-10\,M models are excluded for declaring test-data
leakage in their own leaderboard metadata: Lag-Llama, Super-Linear and the R1 and R2 releases of
TTM. Seven further models are excluded because their size cannot be established. CHARM, DeOS, Lingjiang, LongSeer, Migas and VISIT release neither a checkpoint nor a parameter count. Xihe \citep{xihe} publishes a family from 9.5\,M to 1.5\,B without a per-configuration result that would identify which member the board entry is. Since Xihe and VISIT carry two board directories each, these seven models occupy nine entries. Fifteen further entries are agentic or ensemble systems that route between or combine several forecasters rather than single models, and none publishes a parameter count. Three more are fitted to the evaluation series and one publishes no per-configuration result. FLAIR \citep{flair} sits below the cut and is not a zero-shot model, since it fits a few dozen
coefficients per series at inference, so it is reported with the
statistical baselines rather than in the census. Its three aggregates are the board's own published
values; unlike every other row of Table~\ref{tab:main} they are not recomputed here, because the
snapshot carries no per-configuration file for it. The archived snapshot carries the per-configuration directories the census and the baselines are
computed from.
Chronos-Bolt-Tiny and Chronos-Tiny, which we do use as Chronos-ZS comparators, have no GIFT-Eval
board entry and so cannot enter the census; CITRAS-FM is in the same position.

\paragraph{Parameter counts} Counts come from the released checkpoint where one is available and
from the publishing paper otherwise, and the two do not always agree. Reverso-Small recounts to
\num{550161} learnable parameters against the paper's 550\,K, once $10{,}240$ non-learned FFT
constants are excluded. YingLong-6m holds \num{7319566} parameters despite its release name.
TTM-R3 is listed at its main-branch count. Since the leaderboard run for that entry selects larger-context branches and ensembles across them for $66$ of the $97$ configurations, that count understates what produced its score. TTM-R3 is the entry our $\nWQL$ comparison is measured against.

\paragraph{Per-dataset supervised models} The census is zero-shot only, so models trained on the
series they forecast are out of scope. The board carries ten. The pinned snapshot holds per-configuration
results for eight of them, recomputed here on the same footing as every other comparator
(Table~\ref{tab:supervised}), and TinyCast's $0.774$, $0.545$ and $0.554$ lead all eight on all three
metrics. Of the two the snapshot does not cover, xLSTM-Mixer reports $0.510$ on the board's own
probabilistic metric, ahead of us, and FFM reports $0.704$, behind. That is the comparison the
premise of a pretrained forecaster is meant to displace.

\begin{table}[htpb]
	\centering\footnotesize
	\caption{Per-dataset supervised models in the pinned snapshot, all 97 configurations, ratios to
	seasonal naive. Lower is better. These are trained on the series they forecast and are therefore
	outside the zero-shot census of Table~\ref{tab:main}.}
	\label{tab:supervised}
	\begin{tabular}{@{}lrrr@{}}
		\toprule
		Model            & $\nGMASE\downarrow$ & $\nWQL\downarrow$ & $\nMSIS\downarrow$ \\
		\midrule
		TinyCast (ours)  & 0.774 & 0.545 & 0.554 \\
		\midrule
		PatchTST         & 0.849 & 0.587 & 0.574 \\
		iTransformer     & 0.893 & 0.620 & 0.613 \\
		TFT              & 0.915 & 0.605 & 0.656 \\
		N-BEATS          & 0.938 & 0.816 & 2.512 \\
		DLinear          & 1.061 & 0.846 & 2.841 \\
		TiDE             & 1.091 & 0.772 & 0.906 \\
		DeepAR           & 1.343 & 0.853 & 0.933 \\
		Crossformer      & 2.574 & 1.637 & 6.892 \\
		\bottomrule
	\end{tabular}
\end{table}

\subsection{Comparator intervals}
\label{app:compci}

Table~\ref{tab:main} reports point estimates. Every comparator whose per-configuration results are in
the released snapshot is scored on the same $97$ configurations against the same seasonal-naive
reference, so each difference admits a paired bootstrap, and Table~\ref{tab:compci} gives them. The
delta is our score minus the comparator's, so a positive delta means the comparator leads;
$20{,}000$ replicates, percentile intervals. The second interval resamples the $28$ base datasets
rather than the $97$ configurations, on the grounds of Appendix~\ref{app:protocol}. FLAIR has no
per-configuration entry in the snapshot and so has no interval here. The released artifact carries
the same two intervals for every snapshot model with a per-configuration record, $28$ models and
$84$ deltas, of which $15$ span zero under clustering. The $17$ beyond Table~\ref{tab:main} are the
per-dataset supervised models, the naive baseline, and entries the 10\,M cut put out of range.

\begin{table}[!t]
	\centering\footnotesize
	\setlength{\tabcolsep}{4pt}
	\caption{Paired bootstrap intervals for every difference in Table~\ref{tab:main} with a
		per-configuration comparator record. Positive $\Delta$ means the comparator scores lower
		and therefore leads. The three StatsForecast baselines are omitted: all are resolved against us, by margins of $0.30$ to $8.08$ units. FLAIR is absent for the reason given above, having no per-configuration entry in the snapshot. $\dagger$ marks a difference whose cluster interval spans zero.
		The Reverso rows on $\nWQL$ and $\nMSIS$ carry the same caveat as in
		Table~\ref{tab:main}: those models emit no predictive distribution, so the quantity
		differenced is a point error.}
	\label{tab:compci}
	\begin{tabular}{@{}cllrrr@{}}
		\toprule
		& & & & \multicolumn{2}{c@{}}{95\% CI} \\
		\cmidrule(l){5-6}
		& Model & Params & $\Delta$ & configs & base datasets \\
		\midrule
		\multirow{8}{*}{\rotatebox[origin=c]{90}{$\nGMASE$}}
		& Reverso-Nano & 200\,K & $+0.0140$\rlap{$^{\dagger}$} & $[+0.0030, +0.0245]$ & $[-0.0022, +0.0283]$ \\
		& Reverso-Small & 550\,K & $+0.0475$ & $[+0.0366, +0.0581]$ & $[+0.0322, +0.0613]$ \\
		& TTM-R3 & 1.4\,M & $+0.0498$ & $[+0.0348, +0.0656]$ & $[+0.0290, +0.0695]$ \\
		& Reverso & 2.6\,M & $+0.0626$ & $[+0.0479, +0.0782]$ & $[+0.0437, +0.0843]$ \\
		& Toto-2.0-4m & 4.1\,M & $+0.0172$\rlap{$^{\dagger}$} & $[-0.0081, +0.0382]$ & $[-0.0102, +0.0448]$ \\
		& YingLong-6m & 7.3\,M & $-0.1064$ & $[-0.1446, -0.0733]$ & $[-0.1702, -0.0530]$ \\
		& FlowState-9.1M & 9.1\,M & $+0.0476$ & $[+0.0327, +0.0627]$ & $[+0.0300, +0.0651]$ \\
		& Kairos-10m & 9.9\,M & $+0.0211$ & $[+0.0008, +0.0383]$ & $[+0.0021, +0.0374]$ \\
		\midrule
		\multirow{8}{*}{\rotatebox[origin=c]{90}{$\nWQL$}}
		& Reverso-Nano & 200\,K & $-0.1156$ & $[-0.1324, -0.1002]$ & $[-0.1408, -0.0922]$ \\
		& Reverso-Small & 550\,K & $-0.0811$ & $[-0.0954, -0.0678]$ & $[-0.1006, -0.0634]$ \\
		& TTM-R3 & 1.4\,M & $+0.0259$ & $[+0.0127, +0.0399]$ & $[+0.0085, +0.0459]$ \\
		& Reverso & 2.6\,M & $-0.0649$ & $[-0.0827, -0.0466]$ & $[-0.0886, -0.0382]$ \\
		& Toto-2.0-4m & 4.1\,M & $+0.0213$\rlap{$^{\dagger}$} & $[+0.0001, +0.0401]$ & $[-0.0036, +0.0457]$ \\
		& YingLong-6m & 7.3\,M & $-0.0636$ & $[-0.0894, -0.0405]$ & $[-0.1031, -0.0260]$ \\
		& FlowState-9.1M & 9.1\,M & $+0.0435$ & $[+0.0304, +0.0582]$ & $[+0.0268, +0.0629]$ \\
		& Kairos-10m & 9.9\,M & $-0.0087$\rlap{$^{\dagger}$} & $[-0.0283, +0.0086]$ & $[-0.0278, +0.0092]$ \\
		\midrule
		\multirow{8}{*}{\rotatebox[origin=c]{90}{$\nMSIS$}}
		& Reverso-Nano & 200\,K & $-1.4807$ & $[-1.6688, -1.3123]$ & $[-1.7867, -1.1911]$ \\
		& Reverso-Small & 550\,K & $-1.3911$ & $[-1.5651, -1.2281]$ & $[-1.6846, -1.1185]$ \\
		& TTM-R3 & 1.4\,M & $+0.0528$ & $[+0.0219, +0.0843]$ & $[+0.0076, +0.0935]$ \\
		& Reverso & 2.6\,M & $-1.3506$ & $[-1.5205, -1.1952]$ & $[-1.6226, -1.0875]$ \\
		& Toto-2.0-4m & 4.1\,M & $+0.0989$ & $[+0.0684, +0.1304]$ & $[+0.0668, +0.1309]$ \\
		& YingLong-6m & 7.3\,M & $+0.0200$\rlap{$^{\dagger}$} & $[-0.0131, +0.0522]$ & $[-0.0300, +0.0579]$ \\
		& FlowState-9.1M & 9.1\,M & $-0.0091$\rlap{$^{\dagger}$} & $[-0.0413, +0.0244]$ & $[-0.0488, +0.0215]$ \\
		& Kairos-10m & 9.9\,M & $-0.2224$ & $[-0.2852, -0.1660]$ & $[-0.2967, -0.1549]$ \\
		\bottomrule
	\end{tabular}
\end{table}

\section{Ablations and negative results}

\subsection{Ablation setup}
\label{app:ablsetup}

Table~\ref{tab:families} gives each family's configuration. The exclusion lists of
Appendix~\ref{app:trainsetup} are applied when GIFT-Eval-Pretrain is cached, so every family
inherits them and none saw a benchmark test base. The deployed configuration differs from the
architecture family's causal phase-binning arm in five fields: FFN expansion, feed-forward tying,
convolution factorization, the future-conv readout and the number of quantile levels.

\begin{table}[htpb]
	\centering\footnotesize
	\caption{Configuration of the three ablation families and of the deployed model. Deltas are
		valid within a family and absolute scores are not comparable across families. $Q$ is the number
		of quantile levels; the single-quantile families are scored on $\nMAD$, and the component
		family runs at about a fifth of the deployed sample budget. Every family
		reads GIFT-Eval-Pretrain and Chronos KernelSynth; the component family adds four locally
		generated length-4096 KernelSynth shards, which its synthetic-family arm replaces with the
		four Reverso-family shards the deployed model trains on.}
	\label{tab:families}
	\begin{tabular}{@{}llrrcl@{}}
		\toprule
		Family & Control & Parameters & Steps & $Q$ & Encoder \\
		\midrule
		Architecture  & dilated-conv base            & \num{340545} & 7{,}500  & 1 & untied, non-separable, FFN $1.5$ \\
		              & + phase binning              & \num{361089} &          &   & \\
		              & + recency gate               & \num{393921} &          &   & \\
		              & substituted-recency arm      & \num{410241} &          &   & \\
		\midrule
		Component     & architecture family's best arm & \num{394441} & 30{,}000 & 9 & untied, non-separable, FFN $1.5$ \\
		              & + future-conv                & \num{445513} &          &   & future-conv stack separable, shared FFN \\
		\midrule
		Optimization  & peak LR $3\times10^{-3}$, 100\,M samples & --  & --       & 1 & effective batch 4096 \\
		\midrule
		Deployed      & --                           & \num{146505} & \num{36621} & 9 & tied, separable, FFN $1.0$ \\
		\bottomrule
	\end{tabular}
\end{table}

\subsection{The ablation tables}
\label{app:abltables}

The architecture and optimization arms emit a single quantile, so their weighted quantile loss
reduces exactly to the median absolute deviation. That column is written $\nMAD$, formed and
normalized like $\nWQL$ but a point error, and not comparable to the $\nWQL$ of a nine-quantile
model.

\paragraph{Intervals for the named arms} On the architecture family's 340\,K, 7{,}500-step line,
substituting a phase-free recency path for the detector costs
$0.131$ $\nGMASE$ with a $95\%$ paired bootstrap interval over configurations of $[0.085, 0.189]$,
and on that line it splits by sampling rate, scoring $1.352$ on hourly configurations against
$0.933$ on daily-or-coarser. That arm carries 410\,K parameters against a 340\,K control, so it measures
the detector and the capacity together; the
capacity-matched retrained comparison is the one Section~\ref{sec:ablations} reports. Phase binning's benefit tracks the same axis, rising from a median $0.021$ on the
$24$ configurations the benchmark assigns no seasonality to $0.219$ on the quarter with the largest
ratio of naive to seasonal-naive MASE. The head-to-head against Reverso-Nano runs the other way:
we beat it on $11$ of those $24$ and on $19$ of the $73$ that declare a cycle.
Causal padding is free within resolution on its own (Table~\ref{tab:abl-arch}) and once phase
binning is present, where it costs $+0.0001$ $\nGMASE$ $[-0.0117, +0.0119]$ and $+0.0095$ $\nMAD$
$[-0.0006, +0.0194]$, each spanning zero. Sign symmetrization returns $0.0079$ $\nGMASE$ $[0.0034, 0.0126]$ and $0.0061$ $\nWQL$
$[0.0022, 0.0099]$. The backtest-selected period arm is worse than the raw Fisher pick on both
metrics, by $0.004$ and $0.005$, inside the bootstrap's resolution and inside the kill band
fixed for it before the run. The deployment interaction residual of Section~\ref{sec:exp} is $0.023$
$\nGMASE$.

A Kalman-smoothed decoder variant is released with the other component arms but kept out of
Table~\ref{tab:abl-comp}. At four times the sample budget of the rest of its family it still did not
reach the control, falling short by more than any arm in the table gains. Since its budget differs
from theirs, its delta could not be read against their common control in any case, and it is
excluded from the multiplicity correction below for the same reason.

\begin{table}[htpb]
	\centering\footnotesize
	\caption{Architecture family, dilated-conv base at 340\,K, 7{,}500 steps. Lower is better on $\nGMASE$ and $\nMAD$, so a negative $\Delta$ is an improvement. Arms are not
		parameter-matched, and each row gives its count. $\nMAD$ is the single-quantile point error defined above.
		$\dagger$ marks a delta whose $95\%$ paired bootstrap interval over configurations spans zero.}
	\label{tab:abl-arch}
	\begin{tabular}{lrrrrr}
		\toprule
		& params & $\nGMASE$ & $\Delta$ & $\nMAD$ & $\Delta$ \\
		\midrule
		detector off, recency path substituted & 410\,K & 1.1537 & $+0.1312$ & 0.9774 & $+0.1160$ \\
		control (dilated-conv base) & 340\,K & 1.0225 & & 0.8614 & \\
		causal padding & 340\,K & 1.0161 & $-0.0064$\rlap{$^{\dagger}$} & 0.8562 & $-0.0052$\rlap{$^{\dagger}$} \\
		+ phase binning & 361\,K & 0.9248 & $-0.0977$ & 0.7791 & $-0.0823$ \\
		+ phase binning + causal & 361\,K & 0.9250 & $-0.0975$ & 0.7886 & $-0.0728$ \\
		+ phase binning + recency gate & 393\,K & 0.9092 & $-0.1133$ & 0.7694 & $-0.0920$ \\
		\bottomrule
	\end{tabular}
\end{table}

\begin{table}[htpb]
	\centering\footnotesize
	\caption{Component family, 394\,K nine-quantile line, 30{,}000 steps. Lower is better on $\nGMASE$ and $\nWQL$,
		so a negative $\Delta$ is an improvement. Arms carry the control's parameter count except
		future-conv, which adds $51$\,K. $\dagger$ marks a delta whose $95\%$
		paired bootstrap interval over configurations spans zero.}
	\label{tab:abl-comp}
	\begin{tabular}{lrrrr}
		\toprule
		& $\nGMASE$ & $\Delta$ & $\nWQL$ & $\Delta$ \\
		\midrule
		control (394\,K line) & 0.8123 & & 0.5699 & \\
		+ future-conv correction & 0.7972 & $-0.0151$ & 0.5614 & $-0.0085$ \\
		+ synthetic-family blend & 0.8060 & $-0.0062$\rlap{$^{\dagger}$} & 0.5669 & $-0.0030$\rlap{$^{\dagger}$} \\
		+ synthetic-family blend (tuned dose) & 0.8052 & $-0.0071$\rlap{$^{\dagger}$} & 0.5623 & $-0.0076$\rlap{$^{\dagger}$} \\
		+ gated committing loss & 0.8100 & $-0.0023$\rlap{$^{\dagger}$} & 0.5690 & $-0.0009$\rlap{$^{\dagger}$} \\
		+ future-conv + synthetic families & 0.7860 & $-0.0263$ & 0.5529 & $-0.0170$ \\
		8 training AR chunks (vs 4) & 0.8132 & $+0.0010$\rlap{$^{\dagger}$} & 0.5648 & $-0.0051$\rlap{$^{\dagger}$} \\
		backtest-selected period (vs raw detection) & 0.8167 & $+0.0044$\rlap{$^{\dagger}$} & 0.5749 & $+0.0050$\rlap{$^{\dagger}$} \\
		MASE-weighted loss & 0.8146 & $+0.0023$\rlap{$^{\dagger}$} & 0.5719 & $+0.0020$\rlap{$^{\dagger}$} \\
		\bottomrule
	\end{tabular}
\end{table}

\begin{table}[htpb]
	\centering\footnotesize
	\caption{Optimization family: learning-rate peak and sample budget. Lower is better, so a negative
		$\Delta$ is an improvement. $\nMAD$ is a point error, as
		in Table~\ref{tab:abl-arch}. $\dagger$ marks a delta whose interval spans zero.}
	\label{tab:abl-opt}
	\begin{tabular}{lrrrr}
		\toprule
		& $\nGMASE$ & $\Delta$ & $\nMAD$ & $\Delta$ \\
		\midrule
		control (peak LR $3\times10^{-3}$, 100\,M samples) & 0.8562 & & 0.7252 & \\
		peak LR 1e-3 & 0.8957 & $+0.0395$ & 0.7460 & $+0.0208$ \\
		peak LR 2e-3 & 0.8665 & $+0.0103$ & 0.7300 & $+0.0048$\rlap{$^{\dagger}$} \\
		peak LR 4e-3 & 0.8745 & $+0.0183$ & 0.7304 & $+0.0052$\rlap{$^{\dagger}$} \\
		budget 50\,M samples & 0.8933 & $+0.0371$ & 0.7479 & $+0.0227$ \\
		budget 150\,M samples & 0.8533 & $-0.0029$\rlap{$^{\dagger}$} & 0.7217 & $-0.0035$\rlap{$^{\dagger}$} \\
		\bottomrule
	\end{tabular}
\end{table}

\paragraph{Interval score in the component family} $\nMSIS$ is meaningful only for the component
family, which has the nine-quantile head; the other two families
emit a single quantile, where the score reduces to a rescaled $\nGMASE$. On that metric the longer
training rollout is not null: it improves the interval score by $0.0297$ ($95\%$ CI $[0.0171, 0.0433]$, and
$[0.0163, 0.0435]$ under clustering), on $58$ of the $97$ configurations, which is the only
component arm resolved on $\nMSIS$ under clustering and roughly $2.6$ times the next largest
such delta. Selection ran on the two metrics the family's table carries, and the arm is null
on both, at $+0.0010$ $\nGMASE$ and $-0.0051$ $\nWQL$. Since $\nMSIS$ sits outside the correction
family described below, the interval-score gain is a single uncorrected delta, and we did not carry
the arm into the deployed configuration.

\paragraph{Canonical-period alignment under clustering} Alignment changes only two of the 97
configurations, both from one base dataset, and its bootstrap
interval touches zero for that reason: a replicate that draws neither configuration returns exactly
zero, which happens in $13.2\%$ of draws. Clustering cannot separate it from zero at all: a
replicate that omits \texttt{bizitobs\_l2c} returns exactly zero, which is $36.0\%$ of draws, and
the two-sided achieved level is $0.72$ on both metrics. Its $0.012$ $\nGMASE$ and $0.011$ $\nWQL$
sit inside the headline $0.774$ and $0.545$. Sign symmetrization is
resolved under the same clustering, at $[0.0029, 0.0144]$ $\nGMASE$ and $[0.0019, 0.0111]$
$\nWQL$, and changes all $97$.

\paragraph{Multiplicity and what the correction covers} The correction family is the $36$ deltas
printed in Tables~\ref{tab:abl-arch}, \ref{tab:abl-comp} and~\ref{tab:abl-opt}, which is every arm
of the three families on the two metric columns its own table carries. Each is a $95\%$ paired
bootstrap interval over the $97$ configurations, resampling configurations rather than runs, and
$18$ of the $36$ span zero. Since thirty-six intervals at $95\%$ will resolve one or two by chance,
the family carries a Benjamini--Hochberg correction \citep{benjaminihochberg} at $q{=}0.05$. It leaves $16$ resolved under
configuration resampling. Clustering, which widens the intervals by a median factor of $1.41$,
leaves $15$ or $16$ before the correction and $14$ after it. Read at its strictest, cluster
interval and cluster correction together, the evidence is $14$ deltas, seven arms on both of their
metrics. Detector removal and phase binning are among them, at $[0.063, 0.220]$ and
$[-0.155, -0.049]$ under clustering. The future-conv correction survives neither, losing one metric
or the other under every combination, and its $\nWQL$ cluster interval closes on zero at the upper
end.

The correction is conservative, since ten of the $36$ are
the $\nMAD$ column of a single-quantile arm, where that metric is a point error tracking $\nGMASE$
almost exactly (Pearson $0.994$ across those arms). Thirty-six therefore overstates how many
separate questions were asked, and dropping that column changes no delta's status under either
scheme. The correction also covers the ablation tables and nothing else. The comparator differences
of Appendix~\ref{app:compci}, the substitution arms, the capacity and inference-strategy probes,
sampled feedback and the subgroup splits each carry one interval and no correction, so each is a
$95\%$ statement on its own. The intervals are released with the per-configuration results.

\subsection{Detector evidence}
\label{app:abldetector}

The detector fires on $84\%$ of initial contexts
across all $97$ configurations and fills $2.9$ of its four slots on average, measured over at most
$200$ windows per configuration. Since the deployed model re-runs the detector on every
rolled-forward context, this figure covers the initial context only. Where the true period is
checkable the returned one is right: $24$ for hourly electricity and solar, and $98$, $146$ and
$293$ for the daily cycle at fifteen-minute, ten-minute and five-minute sampling. The small offsets
are bin quantization. Spectral leakage cannot account for the extra slots, since the detector keeps
only local maxima and two adjacent bins cannot both be one. Harmonics account for part of them.
Counting a slot as a new cycle only when it is not a harmonic of one already accepted, a multi-slot
emission carries $2.2$ distinct cycles on average, and $79\%$ carry a second. Raising $K$ therefore
has little left to find, since only $5\%$ reach four distinct cycles and later slots would carry
fewer. A single-delta $K{=}8$ arm on the architecture family's best base bears this out, scoring
$0.9233$ $\nGMASE$ against its $0.9092$ (Table~\ref{tab:overrides}), so we did not carry a larger
$K$ into the deployed configuration.

\paragraph{Firing rate against a frequency control} Accuracy is worse where the detector rarely
fires, and the gap survives a control for declared seasonality. The deployed model scores $0.753$
$\nGMASE$ on the $71$ configurations that fire on more than $90\%$ of windows, against $0.841$ on
the $10$ that fire below $50\%$. Seven of those ten are configurations the benchmark assigns no
seasonality, so the split could be reporting nothing more than that series without a declared cycle
are harder. That reading requires declared seasonality to predict our error, and it does not. Its
rank correlation with $\nGMASE$ is $+0.089$, and the $24$ configurations with no declared cycle
score $0.766$ against $0.776$ for the other $73$. Restricting to those $73$ and splitting them at
$50\%$ firing therefore leaves the gap almost intact, at $+0.074$ $[+0.024, +0.125]$. Within the
architecture family the uncontrolled split moves phase binning from $0.128$ on the $71$ above
$90\%$ to nothing separable from zero on the $10$ below $50\%$, and that null sits with the seven
of those ten the benchmark assigns no seasonality. Only three of the $73$ with a declared cycle
fire below $50\%$, however, so the controlled split's low bucket rests on three points and its
bootstrap admits only ten distinct resamples. Across all $97$ configurations the rank correlation between firing
rate and $\nGMASE$ is $-0.110$ $[-0.290, +0.083]$. The association is in the direction the premise
predicts, and it lives at the tail of the firing distribution rather than across it. The
seasonality contrast of Appendix~\ref{app:abltables} states the question on a larger split.

\paragraph{The detector retrain against each shipped seed} Section~\ref{sec:ablations} contrasts the
detector-off retrain with the mean of three shipped-recipe runs. That contrast is $+0.0071$
$\nGMASE$, $[-0.0004, +0.0148]$ resampling the 97 configurations and $[-0.0013, +0.0177]$ resampling
the 28 base datasets, and $+0.0042$ $\nWQL$, $[-0.0006, +0.0088]$ and $[-0.0005, +0.0089]$. Paired
against each shipped seed separately the contrasts are $+0.0076$ $\nGMASE$
$[+0.0002, +0.0157]$ at seed $42$, $+0.0069$
$[-0.0010, +0.0151]$ at seed $43$ and $+0.0067$ $[-0.0016, +0.0156]$ at seed $44$; on $\nWQL$ they
are $+0.0028$ $[-0.0036, +0.0087]$, $+0.0047$ $[-0.0014, +0.0111]$ and $+0.0051$
$[-0.0008, +0.0111]$.

\paragraph{Substitution arms} The three interventions Section~\ref{sec:ablations} describes score as
follows. Suppressing the detector's output costs $0.0751$ $\nGMASE$ ($95\%$ CI $[0.0440, 0.1138]$).
Against that control the fixed data-blind set is worse still, $0.8597$ against $0.8489$ ($+0.0108$,
$[0.0022, 0.0204]$), and the shuffled draw is barely better, $0.8414$ ($-0.0074$,
$[-0.0141, -0.0001]$), so the detector's own distribution applied to the wrong series recovers almost
none of its value. These interventions leave the phase machinery in
place and change only its input, so they are not comparable with the arm of
Table~\ref{tab:abl-arch} that removes both on a different training line.

\paragraph{Canonical substitution and window occupancy} Substituting the metadata-declared canonical
period for the detector's own choice costs accuracy everywhere, but very unevenly: $0.136$ in log
ratio on sub-hourly and hourly configurations against $0.005$ on daily-or-coarser ones, a contrast
of $0.131$ with a paired bootstrap interval of $[0.084, 0.180]$. The canonical period is the true
seasonal period in both groups, so correctness does not explain the gradient. Window occupancy accounts for it. The substituted period is $288$ samples at five-minute sampling and $360$ at ten-second sampling, so a $2048$-sample context folds onto seven cycles and six. At daily-or-coarser sampling the canonical weekly period of seven packs nearly three hundred cycles into the same window. What the encoder responds to is how many cycles the fold puts in each bin, and the detector
earns its place by choosing periods that both occupy the window well and belong to the series in
front of it. Taken as a whole the canonical substitution is about as damaging as supplying no periods
at all, $0.8509$ against $0.8489$.

Occupancy is not the whole account, and the substitution arms bound how much of the detector's value
it absorbs. Against the $0.0751$ $\nGMASE$ that suppressing the detector's output costs, replacing its choices with a fixed well-spread set costs $0.0859$, more than suppression itself. Its own outputs assigned to the wrong series cost $0.0677$, nine tenths of the way. A well-occupied period the series does not have is therefore worse than none at all, and the part the word ``detection'' most naturally suggests accounts for most of the effect. That the shuffled arm nonetheless beats the
fixed one by $0.018$ $[0.013, 0.024]$ says the distribution of periods the detector produces carries
information about the corpus even when the correspondence to individual series is destroyed.

\subsection{Capacity, size and single-setting sweeps}
\label{app:ablcapacity}

The architecture family bounds how much of phase binning's effect capacity can explain. The recency gate adds $33$\,K parameters to the same base under the same recipe and budget and returns $0.0156$ $\nGMASE$ ($95\%$ CI $[-0.0255, -0.0067]$ over configurations, $[-0.0255, -0.0060]$ under clustering). The next most efficient use of capacity we measured therefore buys about a tenth as much per parameter as phase binning does. Width is measured directly, but on the component
family's \num{394441} line and not on this one: widening $D$ from $64$ to $76$ there, $19\%$ wider
for $41\%$ more parameters, moves $\nGMASE$ by $-0.0018$ ($95\%$ CI $[-0.0081, +0.0047]$) and
$\nWQL$ by $-0.0019$ ($[-0.0063, +0.0021]$), both intervals containing zero.

\paragraph{The reduction to the deployed size} Halving the model cost nothing measurable. Against the \num{445513}-parameter future-conv arm, a
variant that cuts the FFN expansion from $1.5$ to $1.0$, ties the feed-forward, removes the recency
bins and disables the gate reaches \num{225865} parameters, $49\%$ fewer. Paired over the $43$
configurations both arms cover at step $7{,}000$, it scores $1.0085\times$ the control on point
accuracy ($95\%$ CI $[0.9697, 1.0450]$), $1.0085\times$ on probabilistic accuracy
($[0.9892, 1.0321]$) and $0.9531\times$ on the interval score ($[0.9145, 0.9926]$). Both accuracy
intervals contain parity and the interval score improves. The probe covers $43$ of the $97$ configurations and stops well short of the deployed schedule, at
step $7{,}000$. The remaining step to \num{146505}, the separable convolutions, is not separately
paired.

\paragraph{Single-variable overrides} Seven settings were varied one at a time on the architecture
family's best arm, the last row of Table~\ref{tab:abl-arch}, each scored on all $97$ GIFT-Eval
configurations. Table~\ref{tab:overrides} gives them. Six either cost accuracy or sit inside the
family's noise, which is why the shipped values stand. Doubling the phase bins to $32$ is the only setting that moves the other way, by $0.0045$, less than the causal-padding delta the same family's bootstrap cannot separate from zero. That run also postdates the freeze of the deployed line, and $n_b{=}32$ doubles the cyclic template the device holds. We report it without adopting it. Raising the period
cap from four to eight is the
second-largest loss in the table, behind only the recency-weighted fold, and is the evidence behind
stopping at four.

\begin{table}[t]
	\centering
	\caption{Single-variable overrides on the architecture family's best arm. All $97$ GIFT-Eval
	configurations; a negative $\Delta$ is an improvement.}
	\label{tab:overrides}
	\begin{tabular}{lrr}
		\toprule
		& $\nGMASE$ & $\Delta$ \\
		\midrule
		control (phase binning + recency gate) & 0.9092 &            \\
		phase bins, $16$ to $32$              & 0.9047 & $-0.0045$ \\
		cross-horizon convolution, width $5$   & 0.9085 & $-0.0007$ \\
		encoder kernel width, $3$ to $5$      & 0.9100 & $+0.0008$ \\
		gated encoder convolution              & 0.9178 & $+0.0086$ \\
		harmonics per period, $1$ to $2$      & 0.9178 & $+0.0086$ \\
		period cap, $4$ to $8$                & 0.9233 & $+0.0141$ \\
		recency-weighted phase fold            & 0.9242 & $+0.0150$ \\
		\bottomrule
	\end{tabular}
\end{table}
\subsection{Negative results}
\label{app:negative}

Table~\ref{tab:negative} lists the interventions aimed at point accuracy that were tested and
rejected. Each was a single-run probe at reduced budget, so they characterize this architecture at
this scale and need not generalize.

\begin{table}[t]
	\centering\footnotesize
	\caption{Interventions aimed at point accuracy, tested and rejected. $\nGMASE$ throughout, lower
		is better. Groups 1 and 2 are scored against an archived control and group 1's
		per-configuration results are released. Group 3's arms have no archived control, and groups 4
		to 6 are reported by direction.}
	\label{tab:negative}
	\begin{tabular}{@{}lllr@{}}
		\toprule
		Group & Intervention & Scored on & Result \\
		\midrule
		1 Computed values      & seasonal-naive draft as input value & 6-config probe & 0.8825 \\
		                       & + all significant periods           & 6-config probe & 0.8930 \\
		                       & + linear trend term                 & 6-config probe & 0.9178 \\
		                       & control (dilated-conv base)                      & 6-config probe & 0.8760 \\
		                       & trend-seasonal decomposition as channels & 97 configs & 0.9177 \\
		\midrule
		2 Learned rule         & learned per-period weight, variant 1 & 97 configs & 0.9114 \\
		                       & learned per-period weight, variant 2 & 97 configs & 0.9190 \\
		                       & hard reliability rule         & 97 configs & 0.9002 \\
		                       & no rule                       & 97 configs & 0.9092 \\
		\midrule
		3 Per-instance adaptation & 4 methods, best is an oracle gate & 97 configs & 0.818 \\
		                       & amortized model               & 97 configs & 0.774 \\
		\midrule
		4 Tail-robust objective & worst-$\alpha$ per-sample losses     & probe & worse at every $\alpha$ \\
		5 Horizon-weighted loss & linear long-horizon up-weighting     & probe & no net change \\
		6 Median up-weighting   & $q_{0.5}$ term up-weighted           & probe & no gain at convergence \\
		\bottomrule
	\end{tabular}
\end{table}

Group 1 supplied computed quantities as input values rather
than as a coordinate, and the regression grows with the amount of prior supplied. Read together with
the output-suppression arm of Section~\ref{sec:ablations}, this locates what computing buys: a
computed period is useful as a coordinate the encoder is indexed by, while a computed forecast is not
useful as a value the encoder must correct, because the convolutional stack models level and
seasonality in value space better than the hand-computed baselines do. Group 2 tried the premise in
the other direction, replacing the hard reliability rule that rejects periods with too few cycles in
the window by a learned per-period weight. Both variants were worse than the hard rule and worse than
no rule at all, and neither the rule nor its learned replacement is in the deployed configuration.

Training budget interacts with that second result. In a separate probe the hard rule, together with a
wider fixed cross-horizon convolution, improved this family at a reduced training budget and cost
accuracy at the budget the shipped model uses, each flag worth $0.009$ at $30{,}000$ steps against a joint
cost of $0.0161$ at $150{,}000$ steps. The two flags were changed together, so the attribution is joint, but the
direction is what one would expect if a hard-coded bias helps an undertrained model and constrains a
well-trained one.

Group 3 shows the apparent
in-context-learning headroom to be largely a measurement artifact, since a per-series Bayesian fit
that looks strong under a balanced interior-window probe falls behind the amortized model under the
exact GIFT-Eval protocol, and the oracle gate's per-series advantage does not correlate with the
tested identifiability statistics. In group 4 sample-level CVaR \citep{rockafellar2000} concentrates
on the noisiest, most aleatoric samples instead of the under-served configurations, so the
training-sample tail does not transfer to the evaluation-configuration tail the metric scores.

\section{Additional benchmark measurements}
\label{app:extra}

This appendix carries in full the measurements Section~\ref{sec:exp} states in summary.

\subsection{Interval quality and coverage}

$\nWQL$ scores sharpness and calibration
together, and two further measurements separate
them. The first is the mean scaled interval score, normalized the same way and reported in
Table~\ref{tab:main}: TinyCast reaches $0.554$. GluonTS scores it at $\alpha{=}0.05$, so
the $2.5$ and $97.5$ percent levels a nine-decile head does not emit are supplied by the tail
extrapolation of its quantile accessor; every model in the column that emits deciles is treated
alike, but the level is not one any of them predicts directly. Table~\ref{tab:msis} places it against the census. Two models are ahead by a
resolved margin and each spends ten to thirty times the budget, while two further comparisons
do not resolve. FLAIR pretrains nothing and fits $28$ to $57$ coefficients per series, and it has no
per-configuration record, so no interval can be formed against it. The Reverso family emits no
predictive distribution, so its entries are point errors scored by an interval metric
rather than comparable measurements. Those models lead us on point accuracy. Concurrent work adds a quantile head to the
$2.6$\,M Reverso backbone, reported at $3$\,M with the head, and reaches $0.499$ $\nWQL$
\citep{reverso}, twenty times our parameter count and above the $1.4$\,M of the frontier claim.

\begin{table}[t]
	\centering\footnotesize
	\caption{Interval score on GIFT-Eval at the host profile, all $97$ configurations, normalized to
		seasonal naive.
		Lower is better. Positive $\Delta$ means the comparator scores lower and therefore leads.
		Intervals are paired bootstraps over configurations, and $\dagger$ marks a difference whose
		cluster interval spans zero. Parenthesized values are point errors for models not emitting
		predictive distributions, as in Table~\ref{tab:main}, and their differences carry the same
		caveat.}
	\label{tab:msis}
	\begin{tabular}{@{}lrrrl@{}}
		\toprule
		Model & Params~$\downarrow$ & $\nMSIS\downarrow$ & $\Delta$ & $95\%$ CI \\
		\midrule
		TinyCast (ours) & 146\,K          & 0.554\phantom{)}          &                              & \\
		Reverso-Nano    & 200\,K          & (2.035)                   & $-1.4807$                    & $[-1.6688, -1.3123]$ \\
		Reverso-Small   & 550\,K          & (1.945)                   & $-1.3911$                    & $[-1.5651, -1.2281]$ \\
		TTM-R3          & 1.4\,M          & 0.501\phantom{)}          & $+0.0528$                    & $[+0.0219, +0.0843]$ \\
		Reverso         & 2.6\,M          & (1.905)                   & $-1.3506$                    & $[-1.5205, -1.1952]$ \\
		Toto-2.0-4m     & 4.1\,M          & 0.455\phantom{)}          & $+0.0989$                    & $[+0.0684, +0.1304]$ \\
		YingLong-6m     & 7.3\,M          & 0.534\phantom{)}          & $+0.0200$\rlap{$^{\dagger}$} & $[-0.0131, +0.0522]$ \\
		FlowState-9.1M  & 9.1\,M          & 0.563\phantom{)}          & $-0.0091$\rlap{$^{\dagger}$} & $[-0.0413, +0.0244]$ \\
		Kairos-10m      & 9.9\,M          & 0.776\phantom{)}          & $-0.2224$                    & $[-0.2852, -0.1660]$ \\
		\midrule
		FLAIR           & 0\phantom{\,M}  & 0.538\phantom{)}          & --                           & no per-configuration record \\
		\bottomrule
	\end{tabular}
\end{table}

 The second is empirical coverage, computed
from the emitted quantiles over all 97 configurations, weighted by scored points so that a
configuration contributes in proportion to its horizon. Pooled, the
nominal $80\%$ central interval captures $68.0\%$ of actuals, so the model is over-confident. The
deficit tracks the horizon: at the short term the same interval captures $83.0\%$,
at medium $67.4\%$, and at long $60.0\%$. This is the behavior the block-autoregressive decoder
predicts, since every block after the first is conditioned on the median of its predecessors and
the intervals therefore do not widen as they should with accumulated uncertainty. The three terms
draw on different configurations, so the gradient could be composition rather than horizon; restricting
to the $14$ base datasets that appear at all three terms widens it rather than removing it, from $23.0$
to $25.9$ points between short and long. We report the coverage curves per term and per frequency with the evidence.

\paragraph{Interval score at the firmware profile} Table~\ref{tab:main} reports
$\nMSIS$ for the host profile only. Recomputed over the same 97 configurations, the quantized host
profile reaches $0.5632$ against its matched unquantized reference at $0.5535$, a quantizer cost
of $1.75\%$, and the firmware profile reaches $0.6243$ against the host profile's $0.5541$, a cost of
$12.7\%$. The interval score therefore degrades about twice as fast as $\nGMASE$ ($7.6\%$) and
$\nWQL$ ($6.5\%$) over the same step. Empirical coverage is measured on the host path only, so the deployed
configuration's calibration is bounded by this figure rather than measured directly.

\subsection{Quantization spread}

\begin{figure}[!t]
	\centering
	\includegraphics[width=\linewidth]{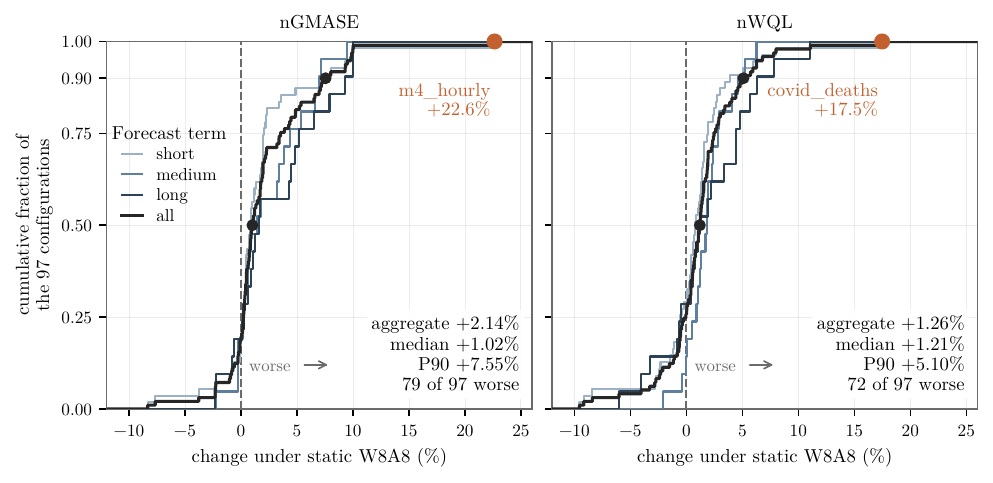}
	\caption{Empirical cumulative distribution over all 97 configurations of the relative change
		from the matched unquantized reference to the quantized host profile, both carrying the two
		host-side strategies, in $\nGMASE$ (left) and $\nWQL$
		(right), for all configurations (black) and by forecast term (light to dark: 55 short, 21
		medium, 21 long). Positive is a degradation and the dashed line is zero change; the panels
		share one $x$ range. Dots mark the median and the P90; the filled marker where a curve
		reaches one is the worst configuration, named with its change.}
	\label{fig:int8scatter}
\end{figure}

\begin{figure}[!t]
	\centering
	\includegraphics[width=\linewidth]{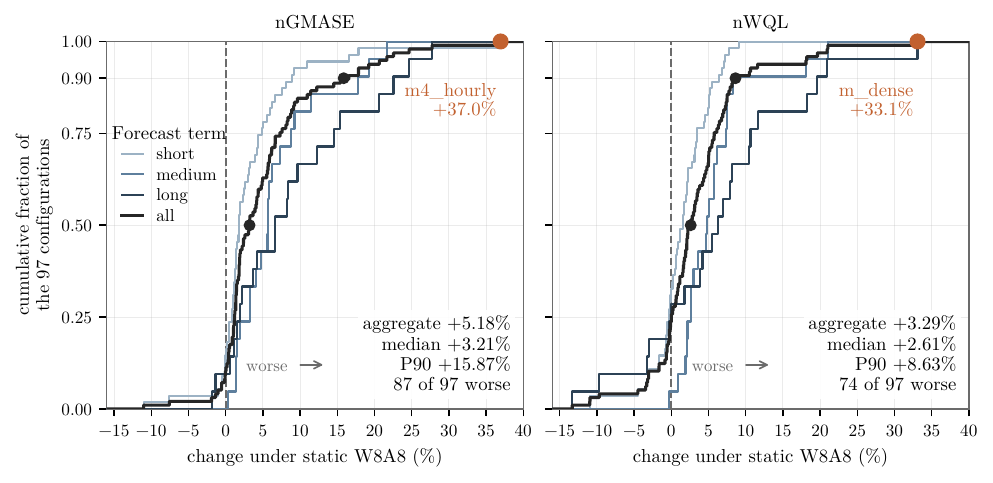}
	\caption{The same construction as Figure~\ref{fig:int8scatter}, but at the firmware profile:
	the relative change from the single-pass reference to exact static W8A8, with
	neither sign symmetrization nor period alignment on either side. Lower is better and positive is
	a degradation.}
	\label{fig:int8scatterdev}
\end{figure}

\paragraph{The quantizer costs more at the firmware profile than at the host profile}
Figure~\ref{fig:int8scatter} pairs the quantizer against a reference that carries both host-side
strategies. The firmware carries neither, and against its own matched single-pass reference the
same quantizer costs more than twice as much: $5.18\%$ $\nGMASE$ and $3.29\%$ $\nWQL$, against
$2.14\%$ and $1.26\%$ at the host profile. The spread widens with it. On $\nGMASE$ the firmware profile degrades $87$ of the
$97$ configurations, against $79$ at the host profile. Its median configuration loses $3.21\%$
against $1.02\%$, its P90 $15.9\%$ against $7.55\%$, and its worst $+37.0\%$ against $+22.6\%$,
in both cases on \texttt{m4\_hourly}, which crosses parity with seasonal naive already at the host
profile, $0.858$ to $1.052$. Nineteen configurations that beat seasonal naive at the host
profile do not at the firmware profile. Figure~\ref{fig:int8scatterdev} gives the distribution.

\subsection{Secondary benchmarks in full}

Chronos-ZS and fev-bench are both scored the way GIFT-Eval is, as geometric means of per-task
ratios to the seasonal-naive reference their own comparator files carry (Tables~\ref{tab:czs} and~\ref{tab:fev}).
Each normalizes over its own task set, so neither is comparable with the other or with the
GIFT-Eval aggregates.

\begin{table}[t]
	\centering\footnotesize
	\caption{Chronos-ZS at the host profile, all $27$ tasks, ratios to the published seasonal-naive
		reference. Lower is
		better. The statistical methods need no training data and have no position on a parameter
		axis.}
	\label{tab:czs}
	\begin{tabular}{@{}lrrr@{}}
		\toprule
		Model & Params & rel.\ MASE & rel.\ WQL \\
		\midrule
		TinyCast          & 146\,K & 0.880 & 0.722 \\
		AutoARIMA         & --              & 0.869 & 0.741 \\
		AutoTheta         & --              & 0.859 & 0.793 \\
		AutoETS           & --              & 0.943 & 0.814 \\
		TTM-R2            & 805\,K          & 1.120 & 1.134 \\
		Chronos-Tiny      & 8.4\,M          & 0.874 & 0.703 \\
		Chronos-Bolt-Tiny & 8.7\,M          & 0.849 & 0.668 \\
		\bottomrule
	\end{tabular}
\end{table}

\begin{table}[t]
	\centering\footnotesize
	\caption{fev-bench at the host profile, all $100$ tasks. Ratios are to the vendored seasonal-naive
		reference and lower is better. Win rate is TinyCast's share of tasks won against that model, computed by the
		benchmark's own pairwise function, so it rests on no comparator pool. AutoARIMA returns $96$
		of the $100$. $^{\ast}$CITRAS-FM releases no checkpoint, so its parameter count is the one
		stated in its paper rather than one we instantiated.}
	\label{tab:fev}
	\begin{tabular}{@{}lrrrrr@{}}
		\toprule
		&                     & \multicolumn{2}{c}{MASE} & \multicolumn{2}{c@{}}{WQL} \\
		\cmidrule(lr){3-4}\cmidrule(l){5-6}
		Model & Params~$\downarrow$ & ratio & win rate & ratio & win rate \\
		\midrule
		TinyCast     & 146\,K                 & 0.819 & --    & 0.658 & --    \\
		AutoARIMA    & 0\phantom{\,M}         & 0.879 & 0.688 & 0.746 & 0.719 \\
		AutoTheta    & 0\phantom{\,M}         & 0.890 & 0.650 & 0.922 & 0.810 \\
		Toto-2.0-4m  & 4.1\,M                 & 0.720 & 0.190 & 0.553 & 0.140 \\
		CITRAS-FM    & 7.2\,M\rlap{$^{\ast}$} & 0.707 & 0.140 & 0.540 & 0.080 \\
		FlowState    & 9.1\,M                 & 0.702 & 0.140 & 0.525 & 0.110 \\
		\bottomrule
	\end{tabular}
\end{table}

Of the three benchmarks, Chronos-ZS is the least favorable to a model built on computed periodicity, and its task mix says why. Of the $27$ tasks, $14$ declare a seasonal period of four steps or fewer and ten declare none at all. The phase fold therefore has little to work with, and extrapolating a trend is the better strategy. On those $14$ we score $0.913$ relative MASE against
AutoARIMA's $0.856$ and AutoTheta's $0.851$; on the other $13$ the ordering inverts and we lead
both, $0.846$ against $0.885$ and $0.867$. The declared period does not account for every task, so
this is a tendency rather than a rule. Against the full field of $21$ published comparators, almost
all of them one to three orders of magnitude larger, TinyCast ranks seventeenth on both metrics.

On fev-bench, disjointness is not established. The pretraining exclusion was defined against the
GIFT-Eval and Chronos-ZS test sets, and twelve fev-bench tasks name corpus subsets we train on. We score $0.934$ relative MASE on those twelve against $0.805$ on the
rest.

We report the ratio-to-seasonal-naive construction used throughout rather than the benchmark's
headline, so every aggregate in this paper is built the same way. The two constructions agree
metric by metric. A skill score is one minus the geometric mean of
the per-task ratios, clipped to $[10^{-2}, 10^{2}]$, and none of our $100$ ratios reaches either
bound, so our relative MASE and WQL give skill scores of exactly $0.181$ and $0.342$. The published
headline $0.304$, with a paired bootstrap over tasks giving $[0.247, 0.364]$, is neither of those:
fev-bench evaluates on a scaled quantile loss, on which our relative error is $0.696$. Its pairwise win rate on that loss is $0.890$ $[0.820, 0.950]$ against seasonal naive.
Table~\ref{tab:fev} gives the rates against each comparator on the two metrics it reports.
We score every task rather than a subset, and omitted tasks are dropped rather than imputed at the
seasonal-naive score. On this benchmark TinyCast leads every statistical baseline on both metrics.

\paragraph{fev-bench with both handicaps removed} The $54$ covariate-free tasks still contain $26$
multivariate ones, which is where a univariate model does relatively best, so that subset controls
only one of the two handicaps the paper names. Among the comparators, CITRAS-FM is the one built
to exploit the inputs at issue, through the cross-variate module of Section~\ref{sec:related}, so
it is where the explanation is most plausible. On the $28$ tasks that are both univariate and
covariate-free, the gap to it is $0.120$ relative MASE and $0.135$ relative WQL, against
$0.113$ and $0.118$ over all $100$ and $0.097$ and $0.101$ on the covariate-free $54$. Removing both
handicaps widens the gap rather than closing it, which is evidence against the reading that the
missing inputs account for it.

\subsection{Inference behavior}

The obvious mitigation for the coverage deficit is to feed a
sampled quantile rather than
the median into the next block, at no change to the inference budget. Measured over all $97$
configurations, its effect falls exactly where the mechanism says it must: $53$ have a horizon no
longer than one block, so nothing is fed back and the two paths are bit-identical. On the $44$ that
recurse the interval score improves in $39$ of them, by $0.087$ $\nMSIS$ with a paired bootstrap
interval of $[0.065, 0.112]$, while point accuracy is worse and $\nWQL$ better by margins the same
bootstrap cannot separate from zero. Pooled coverage rises from $68.0\%$ to $71.7\%$ and the
long-horizon figure from $60.0\%$ to $66.3\%$ with intervals $16\%$ wider, which is under a third of
the distance to nominal, and no intermediate setting buys it more cheaply.

However, two thirds of the long-horizon deficit survives it, so a sampled feedback quantile does
not recover calibration on its own. Whether the residual sits in a quantile head trained too narrow
or in the rollout is not separated here. Taking quantiles across $S$ independent sampled rollouts
would separate them, but it costs $S$ times the inference and falls outside this model's envelope.

\paragraph{Cold start in full} A unit is commissioned before it has history, and the detector
needs a few
cycles in the window before it fires. Truncating the context and rescoring all 97 configurations
gives the curve. The curve is reported on a relative-MAE scale, which is not interchangeable with
$\nGMASE$ and is anchored to this rescoring's own full-context value of $0.759$. The rescoring does
not apply the canonical-period alignment, under which the same checkpoint reaches $0.746$ on the
same scale. Table~\ref{tab:coldstart} gives the curve. The fraction of windows in which the
detector retains a period rises from $25\%$ at $64$ samples to $81\%$ at $512$ and then flattens at
about $84\%$.

\begin{table}[t]
	\centering\footnotesize
	\caption{Cold start: accuracy against the number of observed samples, all $97$ configurations
		rescored on a truncated context at the no-alignment profile, since an alignment factor
		computed from a truncated context would change the effective context length as history
		shrinks. Relative MAE, lower is better; $1.0$ is parity with seasonal naive.}
	\label{tab:coldstart}
	\begin{tabular}{@{}lrrrrrr@{}}
		\toprule
		Observed samples & 64 & 128 & 256 & 512 & 1024 & 2048 \\
		\midrule
		Relative MAE & 1.013 & 0.975 & 0.911 & 0.845 & 0.786 & 0.759 \\
		\bottomrule
	\end{tabular}
\end{table} Accuracy flattens one step later: the step from $512$ to $1024$ samples is still worth $0.058$, and
only the step from $1024$ to $2048$ falls to $0.027$, so the last of the gain is not detector availability. The practical reading is that a fresh deployment degrades
gracefully to baseline instead of failing, and by $512$ samples has recovered two thirds of the
distance from parity to its full-context score.

\subsection{Wins and losses against seasonal naive}

On point accuracy we beat seasonal naive on $90$ of $97$
GIFT-Eval configurations, $22$ of $27$ on Chronos-ZS and $83$ of $100$ on fev-bench.
The losses cluster. Weekly, monthly and annual configurations score $0.873$ $\nGMASE$ against $0.758$ elsewhere. On fev-bench seventeen tasks lose, and the largest margins are \texttt{redset\_5T} at $1.831$, \texttt{proenfo\_gfc14} at $1.439$ and \texttt{redset\_15T} at $1.375$, all of which declare a seasonal period. The monotone trends, \texttt{world\_life\_expectancy} at $1.043$ and \texttt{world\_co2\_emissions} at $1.022$, are among the mildest. On Chronos-ZS they are a near random walk, \texttt{exchange\_rate} at $1.085$, and intermittent retail, \texttt{dominick} at $1.506$. Ten-second GIFT-Eval data at long horizons scores $1.083$. The wins concentrate where two conditions hold. The first is roughly $500$ regular samples, and
the cold-start curve reads $0.911$ at $256$ against $0.845$ at $512$. The second is a cycle the
detector retains, and the $71$ configurations firing above $90\%$ score $0.753$ against $0.841$ on
the $10$ below $50\%$.

A separate limit follows from the head itself. Nine quantiles put the widest central interval at
$80\%$, so $95\%$ and $99\%$ alarm thresholds are unavailable.

\subsection{Qualitative forecasts}
\label{app:pertask}

\begin{figure}[tp]
	\centering
	\includegraphics[width=\linewidth]{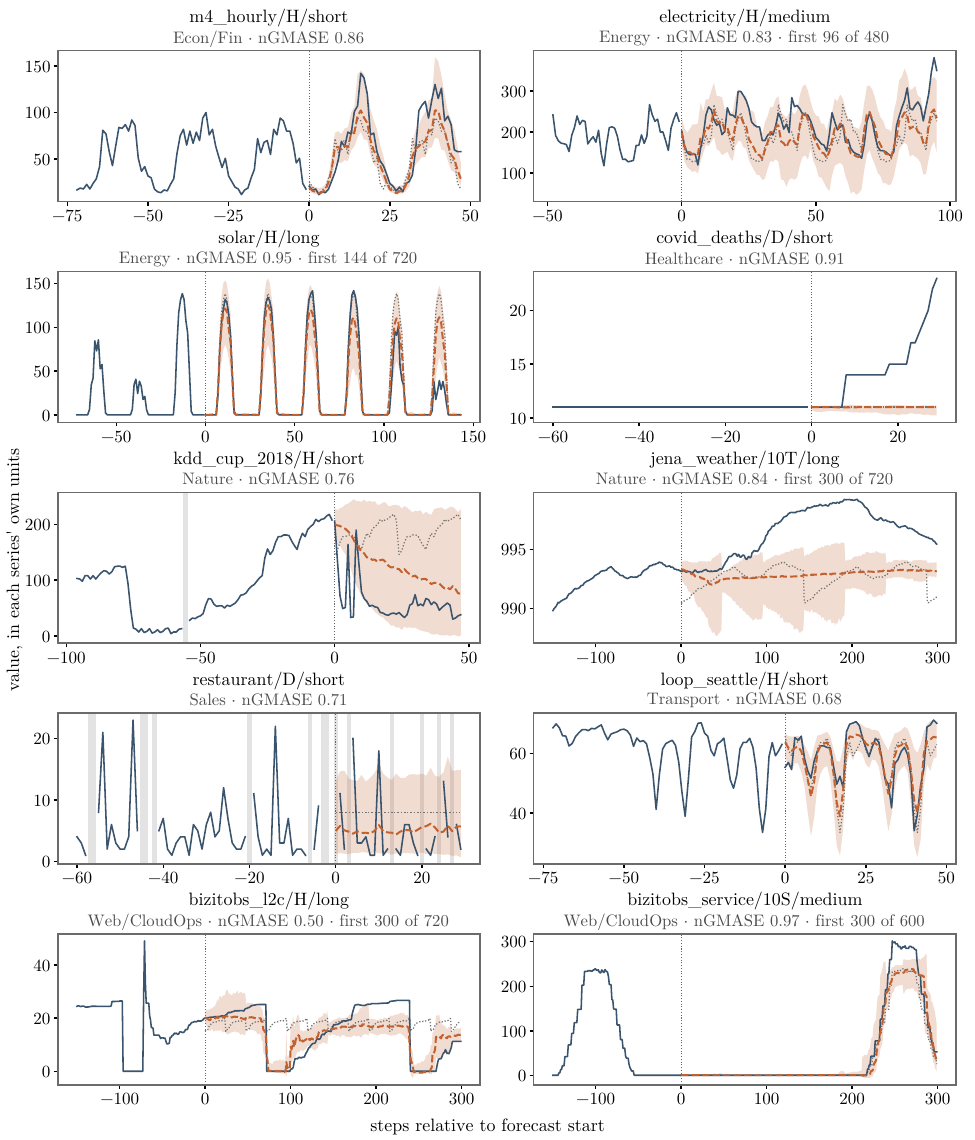}
	\caption{Zero-shot forecasts at the host profile on ten GIFT-Eval tasks, spanning all seven
		domains, four sampling frequencies and three forecast terms. Truth (blue), median forecast (orange,
		dashed), shaded $10$--$90\%$ predictive interval, seasonal naive (grey, dotted); the
		vertical dotted line marks the forecast start and grey bands mark missing observations.
		Titles give configuration, domain, $\nGMASE$ and, where the horizon is only partly drawn,
		``first $n$ of $H$''; $y$ is in each series' own units. The panel-selection and $x$-range rules
		are stated below.}
	\label{fig:qualitative}
\end{figure}
\paragraph{Panel selection for Figure~\ref{fig:qualitative}} Slots are allocated across the seven domains in proportion to their configuration counts, with a floor of one. Within a domain, the configurations at evenly spaced order statistics of its $\nGMASE$ are taken. Within a configuration, the displayed window is the one whose median-forecast MAE is the median over that configuration's windows. The panels are drawn from the same forecasts the reported aggregates come
from. Configurations with a horizon under $12$ steps are excluded from
selection, since a panel of six or eight points cannot show a forecast's behavior, and the
$x$-range is chosen per panel and recorded in the released record. Neither is a choice on score.
The panel median $\nGMASE$ is $0.833$ against a benchmark median of $0.824$, and all $97$
configurations remain in every reported aggregate.

Every per-task figure behind the aggregates of Section~\ref{sec:exp} is in the released record as
CSV, for all three benchmarks, together with the comparator per-task results and the seasonal-naive
denominators. Comparator aggregates are in Table~\ref{tab:main}.

\section{On-device deployment record}
\label{app:deployment}

The figures in Table~\ref{tab:hw} come from a single deployed firmware image on an STM32H753
development board clocked at 480~MHz. Other measurements come from separate builds of the same
runtime, and each is identified where it appears.

\subsection{Setup and operating point}

The model is exported as a static-W8A8 graph: symmetric
per-output-channel weights and per-tensor activations, with all scales frozen before compilation.
The implementation is not integer-only. The min-max normalization and its inverse, every RMSNorm, the SiLU gates with their elementwise products, the bias additions and the period detector evaluate their non-affine arithmetic locally in floating point. The detector works in FP64 and the rest in FP32. There are twenty RMSNorms in the streaming encoder and twelve in the decoder plan, with one more inside the fused decoder FFN-RMSNorm sequence. A hand-written C runtime evaluates the encoder position-by-position with the per-layer ring
buffers of Section~\ref{sec:budget}. The zero-parameter periodicity detector, one rFFT, is treated
as input preprocessing and runs in floating point on the same device. The INT8 matrix multiplications, the
dominant cost, use Arm's CMSIS-NN fully-connected kernels. The phase-mix projection, the fused
decoder FFN and the depthwise convolutions use hand-written integer loops. The image is compiled with GCC at \texttt{-O2} with
$32$-byte function and loop alignment, runs from flash with instruction and data caches enabled, and
copies the encoder weight pack to tightly-coupled memory at start-up. We fix the alignment because
code layout alone moved the measured latency by $1.9\%$ between builds, \SI{4.155}{s} against
\SI{4.076}{s}.

\subsection{Measured performance}

\paragraph{Timing} A core call re-encodes the context position-by-position and decodes one 48-step
block in \SI{4.08}{s}. The position-wise encoder evaluation, at \SI{1.86}{ms} per position, accounts for \SI{3.8}{s} of
the call, and the decoder with its quantile readout for about \SI{0.27}{s} more, of which
\SI{0.20}{s} is the decoder in isolation. Evaluating the encoder position-by-position avoids a full-window activation of about \SI{640}{KiB}, but every call pays the
full window pass. Rebuilding the identical firmware against three matmul backends bounds what kernel
engineering buys: a portable scalar reference takes \SI{6.05}{s}, hand-written SMLAD intrinsics
\SI{4.49}{s}, and CMSIS-NN \SI{4.08}{s}, a $1.49\times$ span, with all three producing identical
outputs. Core-call latency is nearly independent of the input. Across the $32$ contexts of the fidelity
record the entire spread is \SI{9.9}{ms} on a mean of \SI{4.06}{s}, and what variation there
is tracks the number of periodicities the detector accepts, at \SI{1.7}{ms} per active slot. Since the core call performs its own normalization and period detection, the timed region is the
whole forecast. Those $32$ calls were timed on a parity-mode build rather than the canonical
image, so what transfers is the spread and not the level. On the canonical image ten repeated calls
span $0.0025\%$. Longer horizons pay for their blocks and nothing more. Chains of five and fifteen blocks cost five
and fifteen single-call latencies, putting the longest supported horizon of $720$ steps at
\SI{61.1}{s}. The closing quantile sort adds \SI{0.08}{ms} at $H{=}240$ and grows linearly with the
horizon. Since the encoder processes $L$ positions in every case, cold-start forecasts from $1$, $512$ and
$2047$ observed samples cost what a full-context forecast costs.

		\begin{table}[!t]
			\centering\small
			\caption{Deployment record of the TinyCast static-W8A8 core, with FP32 islands, on an
				STM32H753 Cortex-M7 at 480~MHz. The RAM rows are not additive: the first
				\SI{8}{KiB} of the heap row is already reserved in the linker-section row, and the
				accounted total removes the overlap.}
			\label{tab:hw}
			\begin{tabular}{@{}lr@{}}
				\toprule
				\multicolumn{2}{@{}l}{\textit{Latency}}                                \\
				Core call (full $L{=}2048$ re-encode)                & 4.08~s          \\
				Per-position encoder step ($\times L$ per core call) & 1860~$\mu$s     \\
				Decoder (in isolation)                               & 0.20~s          \\
				\midrule
				\multicolumn{2}{@{}l}{\textit{Flash}}                                  \\
				INT8 matrix and convolution coefficients             & 138.1~KiB       \\
				Complete benchmark image (incl. \SI{8}{KiB} context) & 365.5~KiB       \\
				\midrule
				\multicolumn{2}{@{}l}{\textit{RAM}}                                    \\
				Encoder causal rings                                 & 128.5~KiB       \\
				Phase-fold context tensor                            & 128~KiB         \\
				Linker RAM sections and reservation                  & 508.6~KiB       \\
				Persistent model heap payload                        & 217.0~KiB       \\
				Accounted unique RAM (lower bound)                   & 717.6~KiB       \\
				Peak occupancy (statics, heap, stack high-water)     & 730.7~KiB       \\
				\bottomrule
			\end{tabular}
		\end{table}

\paragraph{Storage and memory} The firmware image occupies $17.8\%$ of the device's flash,
including an \SI{8}{KiB} embedded test context. Its INT8 coefficients fall below the
one-byte-per-parameter estimate of \SI{143.1}{KiB}, since the \num{5065} biases and normalization
scalars are carried separately at higher precision. Table~\ref{tab:hw} gives the RAM totals. The
accounted figure is a lower bound, since the heap payload counts model allocations only and excludes
allocator metadata, C-library buffers and stack. Peak occupancy is measured instead. The stack
region is painted at boot and scanned after the run, and the heap break is read back after all
allocations. The measured peak exceeds the accounted bound by exactly the overhead the accounting
leaves out. Two components dominate, and each equals its architectural size exactly at $D{=}64$ and
$L{=}2048$. The encoder causal rings take $\sum_i ((K_c-1)d_i+1)D$ bytes and the phase-fold context
tensor takes $LD$ bytes. Both totals are whole-image figures. The activation arena alone, which is
the quantity embedded inference engines usually report, is \SI{310.1}{KiB}: the \SI{256.5}{KiB} of
rings and phase-fold tensor together with the decoder plan's declared \SI{53.6}{KiB} peak.

\paragraph{Arithmetic and energy} The core call sustains roughly $89$ million multiply-accumulates
per second, one every $5.4$ CPU cycles at 480\,MHz. The distance to the peak rate of the INT8
kernels is spent in the FP32 islands and in the non-matrix operations of the runtime. The part's datasheet
specifies a typical run-mode supply current of \SI{110}{mA} at 480\,MHz on revision-V silicon at
voltage-scaling level VOS0, executing from flash with caches enabled and peripherals disabled, rising
to \SI{148}{mA} from tightly-coupled memory \citep{stm32h753ds}. Our configuration lies between
those rows. At the \SI{3.3}{V} rail those currents bracket $0.36$--$0.49$\,W. At the measured latencies that is an
estimated $1.5$--$2.0$\,J per forecast and $0.7$--$0.9$\,mJ per ingested context position. Counting the
forecast cost alone, a watt-hour of stored energy covers roughly two thousand forecasts. These figures are computed from the datasheet current and the measured latency at
typical silicon, and are not an instrumented measurement.

\subsection{Fidelity}

\paragraph{Calibration protocol} The 32 calibration series are 16 each from Dominick and Wiki
Daily 100K, selected by ranking each source's series on the SHA-256 digest of its identifier, which
reads no sample value. The selection record is released with the paper. Dominick is also a
Chronos-ZS task, so the quantizer's scales are not blind to that benchmark, and it is our worst task
there at $1.506$ relative MASE. Graph inputs take their finite absolute maximum, and internal
thresholds minimize a histogram reconstruction error under a bounded clipping budget. The fixed test
input is excluded. The phase-mix matrix lies outside the compiled plans and is quantized at build
time, so no weight quantization occurs on the device. These frozen scales are what the exact
static-W8A8 rows are computed under. The float simulation in the released code, \texttt{quant.py},
reads an activation scale from each tensor's own range instead, so it does not reproduce those
rows.

\paragraph{Cross-hardware comparison} The chain establishes that the board runs the same
computation as the host, not that it returns identical floats. Board and host are compared on one
fixed test input and on 32 contexts fixed before any board measurement, a different set from the calibration series and sharing
only their count. The contexts are 16 benchmark-workload windows, 8 held-out real series and 8
adversarial inputs. The record was captured on a parity build whose four plan hashes match the
deployed image.

On the fixed test input the board records 432 de-normalized binary32 outputs. Of those, 173 are
bit-identical to the host static-W8A8 reference, and every one lies within 3 units in the last
place, a maximum absolute deviation of $9.5\times10^{-7}$ or about $3\times10^{-7}$ of the output
range. Repeated boots reproduce the record exactly.

Across the 32 contexts, $13{,}824$ outputs in all, period detection is identical everywhere, so both
paths fold the context on the same periods. $55.4\%$ of outputs are bit-identical, and half the
contexts agree within 14 units in the last place throughout. The median relative deviation is zero
pooled and in the held-out real and adversarial tiers. The benchmark-workload tier is $47\%$
bit-identical, so its median is strictly positive.

The largest deviations are discrete rather than accumulated rounding, and they have two sources. A
near-zero output gives a large integer distance at a negligible absolute error. A one-ULP
difference in an FP32 island that falls at an INT8 bin boundary flips the bin, moving that output
by a full quantization step. The worst case is $15.4\%$ of the output range on benchmark-workload contexts,
$16.1\%$ on held-out real series and $22.8\%$ under adversarial stress, the last on a near-constant
input. Three independent INT8 matmul backends, CMSIS-NN, hand-written SMLAD intrinsics and a
portable scalar loop, produce identical outputs on the device, so the integer path itself is
implementation-independent.

\subsection{Limits and projections}

\paragraph{Portability} The streaming encoder, which dominates both latency and state, is not tied
to this device class. Its source and frozen weight pack run unmodified on a commodity \SI{150}{MHz} Cortex-M33
(RP2350), with the same INT8 kernels. There it costs \SI{7.73}{ms} per position within roughly
\SI{230}{KiB}, matching the host scalar reference on the four encoder channels it prints, at $1.30\times$ the
Cortex-M7's cycle count.
The full model exceeds that device's \SI{520}{KB} SRAM, so this measures the portability of the
encoder component rather than a second deployment.

\paragraph{Streaming projection} The deployed firmware re-encodes the whole window. Reusing the same decoder and buffers, a streaming variant would ingest each sample at the measured
\SI{1.86}{ms} per-position step. It would pay the decoder and readout only when a forecast is
emitted, roughly \SI{0.27}{s}, about $15\times$ below the windowed \SI{4.08}{s}. No such firmware was linked or
timed, so these are projected figures.

\end{document}